%% file: top-k-choice.tex
\documentclass[accepted]{uai2022} % after acceptance, for a revised
\usepackage[american]{babel}
\usepackage{natbib} % has a nice set of citation styles and commands
\usepackage{mathtools} % amsmath with fixes and additions
\usepackage{booktabs} % commands to create good-looking tables
\usepackage{tikz} % nice language for creating drawings and diagrams

\usepackage{xcolor}
\usepackage{graphicx, setspace, latexsym,amsmath,amssymb,amsthm,color}

\RequirePackage{etoolbox}
\newtoggle{isdraft}
\relax

\newcommand{\tf}[1]{\boldsymbol{#1}}
\newcommand{\Ah}{\widehat{A}}

\newcommand{\Bh}{\widehat{B}}

\newcommand{\A}{\mathcal{A}}
\newcommand{\B}{\mathcal{B}}

\newcommand{\Ss}{\mathcal{S}}

\newcommand{\Cset}{\ensuremath{\mathcal{C}}}

\newcommand{\R}{\ensuremath{\mathbb{R}}}
\newcommand{\C}{\ensuremath{\mathbb{C}}}

\newcommand{\Loss}{\ensuremath{\mathcal{L}}}

\newcommand{\twonorm}[1]{\lVert #1 \rVert_{2}}
\newcommand{\bigtwonorm}[1]{\left\lVert #1 \right\rVert_{2}}

\newcommand{\mb}[1]{\mathbf{#1}}

\newcommand{\E}{\mathbb{E}}

\renewcommand{\exp}[1]{\mathrm{exp}\left(#1\right)}

\renewcommand{\Pr}{\mathbb{P}}

\newcommand{\PP}{\mathbb{P}}

\newcommand{\Rt}{\tf{\tilde{R}}}
\newcommand{\Mt}{\tf{\tilde{M}}}
\newcommand{\Nt}{\tf{\tilde{N}}}
\newcommand{\Lt}{\tf{\tilde{L}}}

\newcommand{\Rh}{\tf{\hat{R}}}
\newcommand{\Mh}{\tf{\hat{M}}}
\newcommand{\Nh}{\tf{\hat{N}}}
\newcommand{\Lh}{\tf{\hat{L}}}

\newcommand{\hinf}{\mathcal{H}_\infty}

\newcommand{\calS}{\mathcal{S}}
\newcommand{\calE}{\mathcal{E}}

\usepackage{float}

\def\R{\mathbb{R}}

\def\E{\mathbb{E}}

\def\1{\mathbf{1}}

\def\M{\mathbf{M}}

\def\calM{\mathcal{M}}
\def\kl{\mathbb{D}_{\text{KL}}}

\usepackage{subcaption}
\usepackage{algorithm,algorithmic}
\usepackage{pdflscape}
\usepackage{afterpage}
\usepackage{import}
\usepackage[toc,title,page]{appendix}
\usepackage{cancel}
\usepackage{url}

\makeatletter
\newtheorem*{rep@theorem}{\rep@title}
\newcommand{\newreptheorem}[2]{%
\newenvironment{rep#1}[1]{%
 \def\rep@title{#2 \ref{##1}}%
 \begin{rep@theorem}}%
 {\end{rep@theorem}}}
\makeatother

\newtheorem{definition}{Definition}
\newtheorem{theorem}{Theorem}

\numberwithin{theorem}{section}
\newtheorem{lemma}[theorem]{Lemma}

\usepackage{thm-restate}

\newreptheorem{theorem}{Theorem}
\newreptheorem{lemma}{Lemma}
\newreptheorem{coro}{Corollary}

\usepackage{listings}

\title{Efficient and Accurate Top-$K$ Recovery from Choice Data}

\author[1]{\href{mailto:<mdnguyen@seas.upenn.edu>?Subject=Your UAI 2022 paper}{Duc Nguyen}{}}

\affil[1]{%
    Department of Computer and Information Science\\
    University of Pennsylvania.
}
  
\begin{document}
\maketitle

\import{./introduction/}{abstract}
\import{./introduction/}{introduction}

\import{./analysis/}{upper_bound}
\import{./analysis/}{lower_bound}
\import{./analysis/}{connection_mle}

\import{./experiments/}{experiments_compact}

\section{Conclusion}
Ranking under Random Utility Models is a promising area of research with many practical applications. Our work shows how an efficient algorithm can perform very well under a broad family of RUMs. That being said, the class of IID-RUMs constitutes only a subset of models within the class of general RUMs. Beyond IID-RUMs, not much is known in terms of efficient inference and ranking algorithms. In the future, we hope to see more ranking methods developed for more expressive RUMs which have non-identical noise distributions or dependent noise distributions. 

\begin{acknowledgements} The author thanks Shivani Agarwal for suggesting the idea of generalizing the Borda count algorithm to the choice setting, proofreading earlier versions of this paper and for helpful discussions. The author also thanks Prathamesh Patil and William Zhang for proofreading this paper; and the anonymous reviewers for their comments. This work is supported in part by the National Science Foundation (NSF) under grant number 1717290 (awarded to Shivani Agarwal). Any opinions expressed in this paper are those of the author and do not necessarily reflect the views of the National Science Foundation.
\end{acknowledgements}

%%% Bibliography
\bibliography{nguyen_28}

\onecolumn
\import{./proofs/}{proofs}

\import{./experiments/}{experiments_long}
% \bibliography{nguyen_29-supp}

\end{document}

%% file: introduction/abstract.tex
\begin{abstract}
The intersection of learning to rank and choice modeling is an active area of research with applications in e-commerce, information retrieval and the social sciences. In some applications such as recommendation systems, the statistician is primarily interested in recovering the set of the top ranked items from a large pool of items as efficiently as possible using passively collected \emph{discrete choice data}, i.e., the user picks one item from a set of multiple items. Motivated by this practical consideration, we propose \emph{the choice-based Borda count algorithm} as a fast and accurate ranking algorithm for \emph{top $K$-recovery} i.e., correctly identifying all of the top $K$ items. We show that the choice-based Borda count algorithm has optimal sample complexity for top-$K$ recovery under a broad class of \emph{random utility models}. We prove that in the limit, the choice-based Borda count algorithm produces the same top-$K$ estimate as the commonly used Maximum Likelihood Estimate method but the former's speed and simplicity brings considerable advantages in practice. Experiments on both synthetic and real datasets show that the counting algorithm is competitive with commonly used ranking algorithms in terms of accuracy while being several orders of magnitude faster.
\end{abstract}

%% file: introduction/introduction.tex
\section{Introduction}\label{sect:intro}

The research on discrete choice modeling and learning to rank has received a lot of interest in recent years thanks to the growing availability of discrete choice data generated by e-commerce platforms, search engines and the social sciences. In the discrete choice setting, when presented with a set of items, also referred to as \emph{menu}, the user picks the most preferred item. Discrete choice data is an intermediate between pairwise comparison data and full ranking data. In many settings such as e-commerce and political surveys, a large quantity of passively collected data is in the form of discrete choice data, e.g., consumers choosing to buy a product when presented with a catalogue of items, voters picking a favorite candidate from a pool of candidates.

In this paper, we focus on the problem of learning to rank using choice data. Specifically, we are interested in the top-$K$ recovery problem, i.e.,  identifying the set of the top $K$ items out of a universe of $n$ items, using \emph{passively collected choice data}. This problem has many useful applications. For example, in e-commerce applications, marketers are interested in finding the set of the best items based on how consumers make purchasing decisions. In the social sciences, political scientists are interested in determining the most preferred candidates among a pool of candidates using survey questionnaires.

To ground our theoretical discussions, we posit that the choice data is generated according to a probabilistic choice model- when presented with a menu of items $S$, the user makes a non-deterministic decision, picking a single item $i$ from $S$ with some probability $p_{i|S}$.
More specifically, we assume our choice model falls within the class of Random Utility Models with Independently and Identically Distributed noise (IID-RUMs), described in detail in section (\ref{sect:preliminaries}). IID-RUMs are an expressive and flexible framework that can be used to model pairwise comparison data, discrete choice data as well as full ranking data. For example, the Multinomial Logit (MNL) model is one of the most commonly used IID-RUMs to model discrete choice data \citep{train2009discrete}.

\underline{\textbf{Our motivation:}} While expressive, random utility models also pose hard computational problems.
For example, many models within the class of IID-RUMs with the few exceptions such as the MNL model do not admit analytical expression for the choice probabilities (while the pairwise comparison probabilities can be evaluated easily), limiting inference to MCMC-based algorithms. However, sampling-based algorithms can be time inefficient when running on large choice datasets with many items and menus. Furthermore, most classical inference algorithms assume a parametric model generating the choice data. In practice, it is often hard to verify if the data comes from a specific parametric model. Therefore, developing efficient ranking algorithms that are robust to model misspecification is of timely interest. 

Motivated by these considerations, we study the generalization of a simple yet powerful counting algorithm for ranking- Borda count - to the discrete choice setting. The Borda count algorithm itself has a long history, dating back to the 18th century and its analysis has been instantiated in various contexts such ranking from pairwise comparisons in \cite{rajkumar2014statistical,shah2017simple}. Our work, however, is the first to study the theoretical guarantees of Borda count in the \emph{discrete choice setting} under a broad class of discrete choice models.

\underline{\textbf{Our contributions:}}
\begin{itemize}
\item In Section \ref{sect:upper-bound} and Section \ref{sect:lower-bound}, we show that the choice-based Borda count algorithm needs $\theta(n\log n)$ samples in order to exactly recover all of the top $K$ items using choice data. We further show that this sample complexity is optimal for a broad class of IID-RUMs. This hinges on a fundamental property shared by many IID-RUMs which we term \emph{Borda consistency}.
\item In Section \ref{sect:menu-size}, we study the effect of the menu size $m$ on the sample complexity for top $K$ recovery. For the special case of the MNL model, which is a common assumption in the ranking literature, we present an asymptotic characterization of the optimal sample complexity for top $K$ recovery in terms of $m$. This bound monotonically decreases, but at a decreasing rate, with $m$. This suggests that there is a benefit to increasing the menu size but such benefit comes with diminishing returns. To the best of our knowledge, this result is the first of its kind in the choice modeling and ranking literature.
\item In Section \ref{sect:connection-mle-borda}, we study the connections between the choice-based Borda count algorithm and two commonly used top-$K$ recovery algorithms: Maximum Likelihood Estimate under MNL assumption (MNL-MLE) and Spectral Ranking \citep{negahban2017rank,maystre2015fast,agarwal2018accelerated}. We prove that the choice-based Borda count algorithm and MNL-MLE produce the same top-$K$ estimate in the limit of infinite data, even if the data has not been generated by an IID-RUM. On the other hand, Spectral Ranking does not in general give the same estimate as the choice-based Borda count algorithm/MNL-MLE even with infinite data.
\item In Section \ref{sect:experiment-compact}, We show through empirical experiments that the choice-based Borda count algorithm is competitive in terms of accuracy with both MNL-MLE and Spectral Ranking while being several orders of magnitude faster. This highlights the advantage of the choice-based Borda count algorithm in applications where the statistician is primarily interested in efficiently and accurately identifying the top items.
\end{itemize}

\subsection{Related works}
Our work falls within the literature on learning to rank under Random Utility Models (RUMs). 
There has been a substantial amount of work on learning to rank under Random Utility Models and mixtures of Random Utility Models using \textit{full ranking data} \citep{parkes2012random, azari2013generalized, azari2013mom, soufiani2014computing, zhao2016learning,zhao2019learning}. Furthermore, most classical ranking methods assume that the data is generated by a well specified RUM. To the best of our knowledge, our paper is the first to propose a method for top-$K$ ranking under a broad class of RUMs using \textit{passively collected choice data alone}.

The related literature on ranking from pairwise comparisons is vast and we can only refer the interested reader to adjacent problems such as \emph{active top-$K$ recovery} from pairwise comparisons \citep{busa2013top,agarwal2017learning,mohajer2017active,falahatgar2017maxing,falahatgar2018limits,heckel2019active}; top-$K$ recovery from pairwise comparisons \citep{chen2015spectral,shah2017simple,chen2019spectral}; top-$K$ recovery from \emph{$m$-wise sorted data} (full rankings among some $m$ items) \citep{jang2017optimal,chen2020top}.

Closest to our work is the analysis of Borda count by \cite{shah2017simple} who showed that it is optimal for top-$K$ recovery from \textit{pairwise comparisons}. Our work complements theirs by showing that the choice-based Borda count is optimal even in the \emph{general choice setting}. To this end, we obtain in Section \ref{sect:upper-bound} sample complexity upper and lower bounds that are \emph{both more general and refined} than those given by \cite{shah2017simple}. We also study in Section \ref{sect:menu-size} the effect of the menu size on the sample complexity. To the best of our knowledge, our paper presents the first asymptotic characterization of the sample complexity for ranking from choice data in terms of the menu size under the very commonly used MNL model. Operating on \emph{m-wise sorted data}, \cite{jang2017optimal} showed that the optimal sample complexity for top-$K$ recovery under the Plackett Luce model\footnote{Within the ranking literature, Plackett-Luce (PL) is a class of distributions over permutations, induced by the IID-RUM with standard Gumbel noise.} scales with $O(\frac{1}{m})$. Our results complement theirs by showing that the sample complexity for top-$K$recovery from \emph{discrete choice data} scales as $O(1+\frac{1}{m})$. Furthermore, the choice-based Borda count algorithm is different from the Spectral-MLE algorithm studied there which is specialized to the Plackett-Luce model.

\section{Notations and problem formulation}\label{sect:preliminaries}

Let there be $n$ items in the universe. Each item $i$ has a \emph{deterministic and hidden} utility, also referred to as partworth, $U_i$ for $i = 1, \ldots, n$. Let us assume the non-degenerate case where no two items have identical partworths. Without loss of generality, we also assume that $U_{\max} = U_1 > U_2 > \ldots > U_n = U_{\min} > 0$. Let $\Ss_K^* = \{1,\ldots,K\}$ denote the set of $K$ items with the highest parthworths. 

Items are presented to the consumer in a set $S$, also referred to as menu, of size at least 2. When $S$ is presented to the consumer, the perceived utility of each item $i \in S$ is the sum of its parthworth and a random noise term: $X_i = U_i + \epsilon_i$ where the $\epsilon_i$'s are independently and identically distributed according to an \emph{unknown} universal noise distribution $D$. The consumer then picks the item $i$ with the highest perceived utility among all the items in $S$. Such a choice model is referred to as a random utility model with independent and identically distributed noise (IID-RUM). In short, a choice model $\rho$ within the class of IID-RUMs is parametrized by a set of partworths $\{U_1,\ldots U_n\}$ and noise distribution $D$. 

As an overload of notation, we will also use $\rho(i|S)$ to denote the probability that a consumer picks item $i$ from menu $S$ under choice model $\rho$. By definition, $\rho(i|S) = \PP(X_i > X_k\,\forall k \in S\backslash \{i\})\,$. For simplicity, we consider a fixed menu size $m$. However, our analysis can be easily extended to account for a mixture of menu sizes. 

A choice sample is a (menu, item) tuple $(S, y)$ where the consumer chooses item $y$ from menu $S$.
A choice dataset is a set of choice samples.
A top-$K$ recovery algorithm takes in a choice dataset and returns an estimate of the top $K$ items, $\hat\Ss_K$. The goal is to exactly recover the top $K$ items and the performance metric of interest is the 0-1 loss: $L_{01}(\hat \Ss_K, \Ss_K^* ) = \mb 1 [\hat \Ss_K = \Ss_K^*]$ \footnote{$\mb 1$ is the indicator function and the equality is with respect to set equality. }.

We emphasize that as opposed to the \emph{top-$K$ ranking} problem, the objective of the top-$K$ recovery prolem is to accurately identify the set of the top $K$ items, while allowing for mis-ranking among these items.

\section{The choice-based Borda count algorithm}\label{sect:algo}

As discussed previously, the general counting approach referred to as Borda count has a long history and has been instantiated in various contexts such as ranking from pairwise comparisons. Here, we instantiate the Borda count approach to the more general discrete choice setting. This is shown in Algorithm \ref{alg:gbc}.

As would be expected, the algorithm essentially tallies the number of observed `wins' by each item and finally ranking the items by their number of wins, returning the top $K$ items. As with other versions of the Borda count approach, the algorithm is simple and easy to implement; and very efficient in practice. This makes the choice-based Borda count algorithm appropriate in settings where the statistician is primarily interested in efficiently and accurately recovering the set of the top items from a large pool of choice data.

\begin{algorithm}[]
\caption{The choice-based Borda count algorithm}
\hspace*{\algorithmicindent}\textbf{Input: }Choice dataset $\B = \{(S_l,y_l)\}_{l=1}^N$\\
\hspace*{\algorithmicindent}\textbf{Output: }Top-$K$ estimate $\hat \Ss_K$\\
\begin{algorithmic}[1]
  \STATE For each item $i = 1,\ldots, n$\\
  \STATE \hspace{1em} Compute the number of times $i$ gets chosen:\\
  \STATE \hspace{2em} $\hat W_i := \sum_{l=1}^N \mathbf{1}[y_l = i]$\\
  \STATE Return the set of $K$ items corresponding to the highest $\hat W_i$'s. Ties are broken arbitrarily.\\
\end{algorithmic}
\label{alg:gbc}
\end{algorithm}

%% file: analysis/upper_bound.tex
\section{Sample complexity bound}\label{sect:upper-bound}

In this section, we present the sample complexity of the choice-based Borda count algorithm for top-$K$ recovery. We first formalize our sampling model in Section \ref{subsect:sampling-model}. In Section \ref{subsect:generalized-borda-score}, we characterize the class of IID-RUMs under which the choice-based Borda count can successfully identify all of the top $K$ items via a theoretical quantity we term the \emph{generalized Borda score}. The main theorems on the sample complexity of the choice-based Borda count algorithm are presented in Section \ref{subsect:sample-complexity-borda}.

\subsection{The Sampling Model} \label{subsect:sampling-model}
Let $\Cset^{(m)}$ be the set of \textit{all menus} of size $m \geq 2$ (i.e., $\lvert C^{(m)}\rvert = {n\choose m}$). Additionally, let $\Cset_i^{(m)}$ be the set of all menus of size $m$ containing item $i$ (i.e., $\lvert \Cset_i^{(m)}\rvert = {n-1\choose m-1} $). We consider a multiple-round uniform sampling model with $R$ rounds of sampling in total. In each round $r=1,\ldots,R$, each menu $S\in \Cset^{(m)}$ is independently offered with probability $p > 0$. Let $\hat\Cset^{(m,r)}$ denote the set of menus of size $m$ that are offered in round $r$.  If offered menu $S$, the user responds with a random choice $y_S^{(r)}$, where
$$ \Pr(y_S^{(r)} = i) = \rho(i|S)\,.$$
It is easy to check that we have, in expectation, $pR{n\choose m}$ samples over $R$ rounds. 

As a practical example, this sampling procedure can be used to design online political surveys. Suppose that there are $R$ voters willing to take part in answering survey questionnaires to determine support for $n$ political candidates. Fix a ballot size $m$. For each voter, each ballot of size $m$ is independently presented to that user with probability $p$. For each ballot, the voter picks one favourite candidate. 

\subsection{The generalized Borda score}\label{subsect:generalized-borda-score}
For each item $i$, define the following theoretical quantity, which we term the \textit{generalized Borda score}:
\begin{equation*}\label{def:tau}
    \tau^{(m)}_i = \frac{1}{{n-1\choose m-1}} \cdot \sum_{S\in \Cset_i^{(m)}} \rho(i|S) \,.
\end{equation*}
Intuitively, the generalized Borda score is the expected probability that an item $i$ is chosen from a menu $S$ where $S$ is uniformly sampled from $C_i^{(m)}$. Note that $\tau_i^{(m)} \in [0,1]$ for all $i\in [n]$. The generalized Borda score is interesting to us because \emph{for a large class of IID-RUMs}, the order among the items with respect to the generalized Borda scores is the same as that with respect to the partworths. Therefore, it suffices to rank the items by their generalized Borda scores to recover the items with the highest partworths. Formally, we can characterize this class of IID-RUMs using a property we term \emph{Borda consistency}.
\begin{definition}\label{def:borda-consistency} An IID-RUM $\rho$ satisfies \textit{Borda consistency} if for any two items $i, j$ and any menu size $m \geq 2$,
$$\tau^{(m)}_i  > \tau^{(m)}_j \Leftrightarrow U_i > U_j \,.$$
\end{definition}
The follow lemma establishes that many commonly used IID-RUMs such as the MNL (Gumbel distributed noise) and the Probit (Normal distributed noise) model satisfy Borda consistency.
\begin{lemma}\label{lem:iid-rum-order} All IID-RUMs whose noise distribution has absolutely continuous density function and support on the real line satisfy Borda consistency.
\end{lemma}

In the supplementary materials, we will show that Borda consistency is satisfied by an even broader class of IID-RUMs that include other commonly used models such as the IID-RUM with exponentially distributed noise. Intuitively, this stems from the property enjoyed by many IID-RUMs: for any two items $i, j$ where $U_i > U_j$, $\rho(i|S) > \rho(j|S) \,\,\forall S \in \Cset^{(m)}:i, j\in S$; and $\rho(i|S\cup \{i\}) > \rho(j|S\cup\{j\}) \,\,\forall S \in \Cset^{(m-1)}: i, j\notin S$. To the best of our knowledge, this fundamental property that holds across a very broad class of IID-RUMs has not been previously decribed in the literature and may be useful to future works exploring the intersection of ranking and choice modeling.

\subsection{Exact top-$K$ recovery }\label{subsect:sample-complexity-borda}
Having established Borda consistency as a property enjoyed by many IID-RUMs, we will now present the finite sample guarantees of the choice-based Borda count algorithm for top-$K$ recovery that holds for all choice models in this broad class of IID-RUMs. To show that the choice-based Borda count algorithm accurately identifies all of the top $K$ items with high probability, it suffices to bound the probability that the algorithm mistakenly ranks an item $j \notin \Ss_K^*$ higher than another item $i \in \Ss_K^*$. Specifically, we want to bound the following probabilities.
$$ \PP(\hat W_j > \hat W_i) \quad \forall i \in \Ss_K^*, j\notin \Ss_K^* \,,$$
where $\hat W_i$ is defined in Algorithm (\ref{alg:gbc}). 
Considering this, the fundamental hardness of top-$K$ ranking lies in distinguishing between the $K$-th and $K+1$-th best item, and therefore depends on the gap between their generalized Borda scores: 
\begin{equation*}\label{def:delta-K}
    \Delta_K^{(m)} = \tau^{(m)}_{K} - \tau^{(m)}_{K+1} \,.
\end{equation*}
The smaller this gap, the more data the algorithm requires in order to correctly separate between the top $K$ and the bottom $n-K$ items. Building on this intuition and generalizing to any pair of items $(i,j)$ where $\tau_i > \tau_j$, we obtain the following upper bound on $ \PP(\hat W_j > \hat W_i)$.
\begin{lemma}\label{thm:error-bound-gen-borda}
Consider an IID-RUM that satisfies Borda consistency per Definition \ref{def:borda-consistency}. Assume input choice data with menu size $m$ is generated according to the sampling model described in Section \ref{subsect:sampling-model}. For any two items $i$ and $j$ where $\tau_i^{(m)} > \tau_j^{(m)}$, the choice-based Borda count algorithm satisfies
\begin{multline*}
\PP(\hat W_j > \hat W_i) \leq \exp{\frac{-3pR{n\choose m} {m(\tau_i^{(m)}-\tau_j^{(m)})}^2 }{8n(\tau_i^{(m)} +\tau_j^{(m)})}} \,.
\end{multline*}
\end{lemma}

The proof of Lemma \ref{thm:error-bound-gen-borda} uses a standard concentration inequality argument based on Bernstein's inequality (cf. Theorem 2.8.4 \cite{vershynin2018high}). The lemma itself states that, for each pair $i\in \Ss_K^*, j\notin \Ss_K^*$, if $pR{n\choose m} \geq  {\frac{8n\log n (\tau_i^{(m)} + \tau_j^{(m)} )}{m(\tau^{(m)}_i - \tau^{(m)}_j)^2}}$, then $\Pr(\hat W_j > \hat W_i) = O(\frac{1}{n^3})$. We also have the following lemma which presents an upper bound on the item-dependent term $ \frac{\tau^{(m)}_i + \tau^{(m)}_j}{(\tau^{(m)}_i - \tau^{(m)}_j)^2 }$.

\begin{lemma}\label{lem:simple-inequality} 
Consider an IID-RUM that satisfies Borda consistency per Definition \ref{def:borda-consistency}. For any $K$, we have
$$ \frac{\tau^{(m)}_K + \tau^{(m)}_{K+1}}{{\Delta_K^{(m)}}^2 } = \max_{i\in \Ss^*_K, j\notin \Ss^*_K} \bigg\{ \frac{\tau^{(m)}_i + \tau^{(m)}_j}{(\tau^{(m)}_i - \tau^{(m)}_j)^2 } \bigg\} \, .$$
\end{lemma}

By combining the two lemmas above and applying union bound over all pairs $i\in \Ss_K^*, j\notin \Ss_K^*$, we obtain the following sample complexity bound for exact top-$K$ recovery:
\begin{theorem}\label{cor:sample-complexity-gbs}
    Assume the conditions of lemma (\ref{thm:error-bound-gen-borda}). Given sufficiently large $p, R$ such that $pR{n \choose m} \geq \frac{8n \log n}{m{\Delta_K^{(m)}}^2}\cdot(\Delta_K^{(m)} +2\tau_{K+1}^{(m)})$, the choice-based Borda count algorithm correctly identifies all of the top $K$ items with probability at least $1-O(\frac{K}{n^2})$.
\end{theorem}
The reader may also recognize that $\Delta^{(m)}_K + 2\tau_{K+1}^{(m)}$ is simply $\tau_K^{(m)}+\tau_{K+1}^{(m)}$. The former presentation is, however, useful in highlighting the main quantities that will also reappear in our matching lower bound. In summary, the choice-based Borda count algorithm has the following sample complexity for exact top-$K$ recovery:
$$O\bigg(\frac{n\log n}{m{\Delta_K^{(m)}}}\cdot(1 + \frac{\tau_{K+1}^{(m)}}{\Delta_K^{(m)}} )\bigg)\,.$$
This shows that overall, we only need $O(n\log n)$ examples to recover the top $K$ items from choice data with high accuracy. Our upper bound (and matching lower bound to be shown) can be seen as both \emph{generalization and refinement} of Theorem 1 of \cite{shah2017simple}. Under the pairwise comparison setting ($m=2$), we can simply upper bound $\tau_{K+1}^{(m)} \leq 1$ and recover the (optimal) sample complexity $O\big(\frac{n\log n}{{\Delta_K^{(2)}}^2}\big)$ of Borda count obtained by \cite{shah2017simple}. The analysis approach there, however, is insufficient to produce an optimal sample complexity bound in the discrete choice setting. Note also that there can be combinatorially many realizations of the data in the discrete choice setting as $\lvert \Cset^{(m)} \rvert = {n\choose m}$. Our proof therefore requires considerably more effort. Our bound also shows that the sample complexity depends not only on the gap $\Delta_K^{(m)}$ between the $K$-th and $K+1$-th item, but also the relative `strength' of the $K+1$-th item, as captured by the $\frac{\tau_{K+1}^{(m)}}{\Delta_K^{(m)}}$ term.

In general, the factors $\tau_{K+1}^{(m)}$ and $\Delta_K^{(m)}$ don't admit closed form expressions because both are sums of ${n-1\choose m-1}$ terms. The reader may also recognize that these parameters also depend on the menu size $m$, the partworth parameters and the noise distribution. In the next section, we will show a \emph{matching lower bound} in terms of the same parameters, establishing the optimality of the choice-based Borda count algorithm, and discuss why the exact relation between $\Delta_K^{(m)}$, $\tau_{K+1}^{(m)}$ and the model parameters remains elusive. 

Often in practice, we may tolerate some error for top-$K$ ranking by allowing the algorithm to misidentify, up to a threshold, some number of items. This is known as \emph{approximate top-$K$ recovery}. We include detailed discussions of this problem in the supplementary materials and show that the choice-based Borda count algorithm also has \emph{optimal sample complexity} for approximate top-$K$ recovery under the broad class of IID-RUMs that satisfy Borda consistency.

%% file: analysis/lower_bound.tex
\section{Information-theoretic lower bound}\label{sect:lower-bound}
In this section, we will show that the choice-based Borda count algorithm enjoys optimal sample complexity by furnishing a matching lower bound. 
To show a lower bound, we will construct a special subclass of the MNL family where any estimator requires $\Omega(n\log n)$ examples in order to exactly recover the top $K$ items. We defer detailed descriptions of this model to the supplementary materials while stating the main results as follows.
\begin{theorem}\label{thm:lower-bound-all} 
Consider the sampling model described in Section \ref{subsect:sampling-model}.
There exists a class of MNL models such that for $n \geq 20$, if $pR{n\choose m} \leq \frac{n\log n }{8}\cdot\frac{\tau_{K+1}^{(m)} +\Delta_K^{(m)} }{m{\Delta_K^{(m)}}^2}$ then any estimator fails to correctly identify all of the top $K$ items with probability at least $\frac{1}{12}$.
\end{theorem}

The proof of Theorem (\ref{thm:lower-bound-all}) first reduces the problem of exact top-$K$ recovery to a multiple hypothesis testing problem and then applies Fano's lemma \citep{cover1999elements}. Each hypothesis in the testing problem corresponds to an MNL model. Within each model, the set of the top $K$ items always includes items $1,\ldots, K-1$. However, the index of remaining item in the top-$K$ set is different for each model (i.e., there are $n-K+1$ different models). We make all of the top $K$ items have the same partworths while the bottom $n-K$ items have the same (and lower) partworths. The key challenge then is to obtain a tight upper bound on the KL divergence between any two hypothesis models. 
In summary, Theorem (\ref{thm:lower-bound-all}) implies the following \textit{minimum} sample complexity for any algorithm for top-$K$ recovery:
$$ \Omega\bigg( \frac{n\log n}{m{\Delta_K^{(m)}}} \cdot \big(1 + \frac{\tau_{K+1}^{(m)}}{\Delta_K^{(m)}}\big) \bigg) \,.$$
Comparing with the bound in Theorem \ref{cor:sample-complexity-gbs}, one can see that the sample complexity of Borda Count is optimal in terms of both $m$, $n$ as well as the model dependent parameters $\Delta_K^{(m)}$ and $\tau_{K+1}^{(m)}$.

\section{The role of the menu size $m$}\label{sect:menu-size}
The effect of the menu size on the performance of top-$K$ recovery algorithms is an aspect of both theoretical and practical importance. In real life applications, the menu size could range from 2 to hundreds of items. One may suspect that increasing the menu size means the data carries more information per data point, and thereby reduces the sample complexity for top-$K$ recovery. However, to the best of our knowledge, such a relationship has not been theoretically established in the literature on choice modeling, even for the very commonly used MNL model.

As seen in the matching lower and upper bound for the sample complexity of top-$K$ recovery, the menu size enters in complex ways through the factors $\Delta_K^{(m)}$ and $\tau_{K+1}^{(m)}$. Both factors can vary in subtle ways with $m$, depending on the underlying choice model. Even for the class of MNL models which admit closed form choice probabilities, these factors don't seem to have a closed form expression as each of them is a sum of ${n-1\choose m-1}$ terms. To bypass the difficulty of exactly evaluating $\Delta_K^{(m)}$ and $\tau_{K+1}^{(m)}$, we characterize the asymptotic dependency of $\frac{1}{m{\Delta_K^{(m)}}}$ and $\frac{\tau_{K+1}^{(m)}}{\Delta_K^{(m)}}$ on $m$ \emph{under the MNL class of models} and show that both of these factors monotonically decrease with $m$ but at a \emph{decreasing rate}. This implies that while there is an advantage to using choice data of larger menu sizes, there is a diminishing return to increasing the menu size.
\begin{theorem}\label{thm:mnl-variational-lower-bound} For any MNL model and a fixed $K$,
    \begin{equation*}
     \frac{1}{m\Delta_K^{(m)}} = \theta\bigg( \frac{1}{e^{U_K}-e^{U_{K+1}}}\cdot \big(1 + \frac{1}{m-1}\big)  \bigg)\,,
    \end{equation*}
    \begin{equation*}
    \frac{\tau^{(m)}_{K+1}}{\Delta_K^{(m)}} = \theta\bigg( \frac{e^{U_{K+1}}}{e^{U_K}-e^{U_{K+1}}}\cdot \big( 1+ \frac{1}{m-1}\big) \bigg) \,.
    \end{equation*}
\end{theorem}
It can be seen in both $ \frac{1}{m\Delta_K^{(m)}}$ and $\frac{\tau^{(m)}_{K+1}}{\Delta_K^{(m)}}$ that the term which depends on $m$, $1 + \frac{1}{m-1}$, montonically decreases with $m$ but at a diminishing rate. Combining the above theorem and the matching sample complexity bounds obtained earlier, one can see that the optimal sample complexity for top-$K$ recovery from choice data scales as $\theta(1+\frac{1}{m})$.

\begin{figure}[]
    \centering
        \includegraphics[scale=0.5]{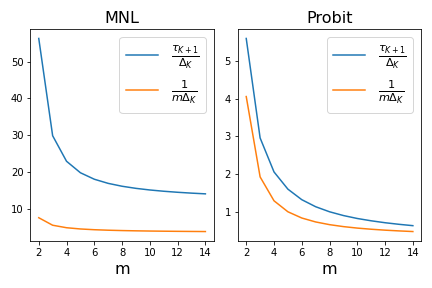}
    \caption{$\frac{1}{m{\Delta_K^{(m)}}}$ and $\frac{\tau_{K+1}^{(m)}}{\Delta_K^{(m)}}$ decrease with larger $m$ under a randomly generated MNL and Probit model. This suggests that there is an advantage, albeit with dimnishing return, to using larger menu sizes. }
    \label{fig:model-quantities}
\end{figure}

Outside of the MNL family of models, we are not aware of any IID-RUM that admits a closed form expression for the choice probabilities. However, suppose that we know all of the partworths and the noise distribution, we can still approximate the choice probabilities via Monte Carlo sampling. Given these (approximated) choice probabilities, one can then evaluate $ \frac{1}{m\Delta_K^{(m)}}$ and $\frac{\tau^{(m)}_{K+1}}{\Delta_K^{(m)}}$. 
As an example, Figure \ref{fig:model-quantities} shows how these quantities vary with $m$ under a randomly generated MNL and Probit model (IID-RUM with standard normal noise) \citep{train2009discrete} with $n=15$, $K=3$. The partworths were independently generated from a zero-mean normal distribution which is also a commonly chosen prior in the literature \citep{parkes2012random, train2009discrete}. The curves for $\frac{1}{m{\Delta_K^{(m)}}}$ and $\frac{\tau_{K+1}^{(m)}}{\Delta_K^{(m)}}$ decrease at a rate approximately similar to those of the MNL model as stated in Theorem \ref{thm:mnl-variational-lower-bound}.
Ranking from choice data under MNL model assumptions remains an active area of research \citep{agarwal2018accelerated,agarwal2020choice} and to the best of our knowledge, our work presents \emph{the first asymptotic characterization} of the optimal sample complexity for top-$K$ recovery in terms of the menu size $m$ under this often used class of choice models.

%% file: analysis/connection_mle.tex
\section{Connections to commonly used ranking algorithms}\label{sect:connection-mle-borda}

In this section, we establish close connections among choice-based Borda count, the method of maximum likelihood estimate under MNL assumptions (MNL-MLE) \citep{train2009discrete} and Spectral Ranking \citep{negahban2017rank,maystre2015fast,agarwal2018accelerated} which will explain many experimental results we present in later sections.

Firstly, one can prove that choice-based Borda count and MNL-MLE are `equivalent' top-$K$ recovery algorithms in the limit of infinite data. This connection is formalized as follows.
\begin{theorem}\label{thm:borda-mle-connection} Consider the sampling model described in Section \ref{subsect:sampling-model}, for any $p > 0$, in the limit as $R\rightarrow \infty$, MNL-MLE and choice-based Borda count will produce the same top-$K$ estimate. Moreover, this holds even if the data does not come from the MNL model or any IID-RUM.
\end{theorem}

A similar observation was made in \cite{rajkumar2014statistical}: under the \emph{pairwise comparison} setting the Borda count algorithm and MNL-MLE are both consistent for full ranking under a class of pairwise comparison models that is strictly more general than the BTL model \footnote{The BTL model is the instantiation of the MNL to the pairwise comparison setting}. Our results generalize the relation between the two algorithms to the choice setting and show that in fact the Borda count algorithm and MNL-MLE produce the same estimate in the limit of infinite data under any choice models. This connection between the two methods is reflected in our experiments where the performance of the choice-based Borda count algorithm is almost identical to that of MLE, when the sample size is large. While performing similarly to MNL-MLE, the choice-based Borda count algorithm is several orders of magnitude faster thanks to its simplicity. This suggests that if the statistician is mostly concerned with recovering a small number of top items, the choice-based Borda count algorithm should be seriously considered due to its speed, simplicity and guaranteed optimal sample complexity.

The above result also means that MNL-MLE is a consistent top-$K$ ranking algorithm under the broad class of IID-RUMs, since the choice-based Borda count algorithm is consistent in recovering the top $K$ items. This shows that MNL-MLE may be used for ranking applications even when the data does not satisfy the MNL assumption. Consistency of MLE under model misspecification is an underexplored question and we leave the careful characterization of the sample complexity of MNL-MLE when the data comes from a non-MNL distribution as a subject of future studies.

On the other hand, Spectral Ranking does not in general produce the same top-$K$ estimate as MNL-MLE/choice-based Borda count. However, when the underlying choice model falls within a broad class of IID-RUMs which include many commonly used choice models such as the MNL and Probit model, all three algorithms produce the same estimate given infinite data.

\begin{theorem}\label{thm:borda-mle-asr-connection} Consider the sampling model described in Section \ref{subsect:sampling-model}. Assume that the underlying choice model generating the data is in the class of IID-RUMs whose noise distribution has absolutely continuous density function with support on the real line. For any $p > 0$, in the limit as $R\rightarrow \infty$, then Spectral Ranking, MNL-MLE and choice-based Borda count produce the same top-$K$ estimate.

On the other hand, there exists a choice model where in the limit as $R\rightarrow \infty$, the Spectral Ranking algorithm produces a different top-$K$ estimate from MNL-MLE/Borda count.
\end{theorem}

%% file: experiments/experiments_compact.tex
\section{Experiments}\label{sect:experiment-compact}
In this section, we present experiment results on both synthetic and real datasets. The main performance metric is top-$K$ accuracy. More specifically, we measure top-$K$ accuracy as the frequency at which the respective algorithm correctly identifies \emph{all} of the true top $K$ items, \emph{over 100 trials}.

\subsection{Synthetic data}\label{subsect:experiments-synthetic}
\begin{figure}[h]
\centering
\hbox{
    \hspace{-.5cm}
    \includegraphics[scale=0.35]{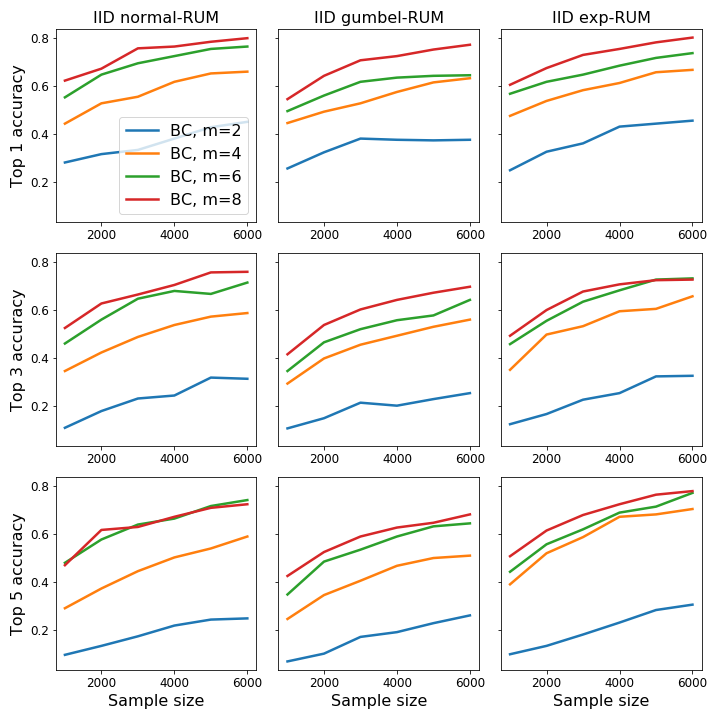}
}
\caption{\textbf{Synthetic data:} Exact top $K$ accuracy of choice-based Borda count against sample size for different menu sizes 2 ({\color{blue}blue}), 4 ({\color{orange}orange}), 6 ({\color{green}green}), 8 ({\color{red}red}) with $K=1,3,5$ and $n=50$. Increasing the menu size improves performance but with a diminishing return.}
\label{fig:synthetic-borda-compact}
\end{figure}

We verify, via synthetic experiments, the efficacy of the choice-based Borda count algorithm and the effect of the menu size $m$ on its performance.
Let there be $n=50$ items in the universe. We experiment with 3 different noise distributions: standard Normal noise (Probit), standard Gumbel noise (MNL) and standard Exponential noise. We vary the menu size $m=2,4,6,8$ and $K = 1,3,5$. Figure \ref{fig:synthetic-borda-compact} shows top-$K$ accuracy against the sample size. In all experiments, choice-based Borda count successfully identifies the top $K$ items with high probability given sufficiently large sample size. Furthermore, using larger menu sizes improves the performance of Borda Count. However, it can be seen that there is a diminishing return in performance gains from using larger menu sizes, agreeing with our theoretical analysis in Section \ref{sect:lower-bound}.

\subsection{Real data}\label{subsect:experiments-real}

\textbf{Baseline algorithms:} We compare choice-based Borda count against Accelerated Spectral Ranking (ASR) \citep{agarwal2018accelerated} and Maximum Likelihood Estimate (MLE) under MNL assumptions \citep{train2009discrete} in terms of top $K$ accuracy. We implement MLE using Scipy's L-BFGS optimizer \citep{2020SciPy}.

\textbf{Data description:} We follow standard procedures commonly used in previous works such as \cite{rajkumar2014statistical,agarwal2020choice}. Operating on full ranking datasets, we can estimate the choice probabilities for any menu $S$, i.e., choice probability $\rho(i|S)$ is the proportion of rankings that ranks item $i$ highest among all them items in $S$. Given these probabilities, we can simulate the sampling model as described in Section \ref{subsect:sampling-model}. Our datasets include SUSHI \citep{kamishima2003nantonac}, APA election dataset \citep{diaconis1989generalization}, 3 Irish election datsets and F1 race dataset included in the library PrefLib \citep{mattei2013preflib}. Notably, the induced pairwise choice probabilities of these datasets all satisfy stochastic transitivity. Therefore, there exists a universal ordering of the items which we can use as a true global ranking over the items. Due to space constraint, we can only present a few representative experimental findings and leave additional results with detailed descriptions of data processing in the supplementary materials. 

\begin{figure}[]
\centering
\hbox{
    \includegraphics[scale=0.40]{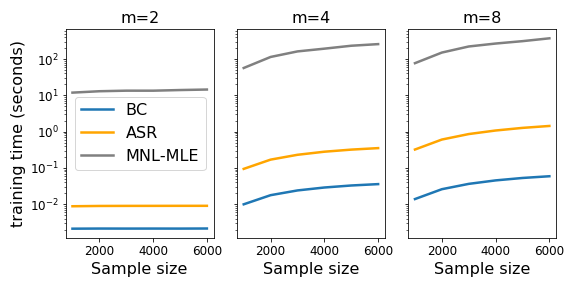}
}
\caption{\textbf{F1 dataset ($n=22$):} Average training time (seconds) against sample size for $m=2,4,8$. Choice-based Borda count ({\color{blue} blue}) is several orders of magnitude faster than its competitors.}
\label{fig:f1-time-compact}
\end{figure}

\textbf{Speed advantage:} Across all experiments, choice-based Borda count is several orders of magnitude faster than ASR and MLE. This difference is especially pronounced in datasets with more items such as the F1 dataset, as shown in Figure \ref{fig:f1-time-compact}.

\begin{figure}[]
\centering
    \includegraphics[scale=0.4,height=80mm]{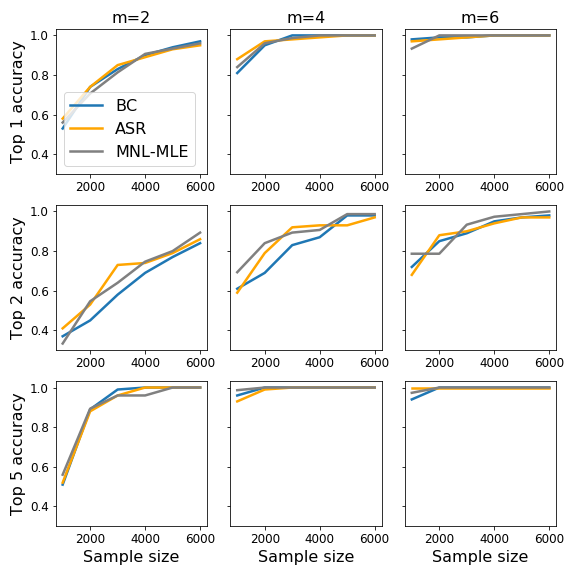}
\caption{\textbf{Irish-Meath dataset ($n=14$):} Exact top-$K$ accuracy against sample size. choice-based Borda count ({\color{blue}blue}) is competitive with baseline algorithms. Using larger menu sizes generally improves the performance of the algorithms.}
\label{fig:irish-meath-experiment-compact}
\end{figure}

\begin{figure}[]
\centering
    \includegraphics[scale=0.4,height=80mm]{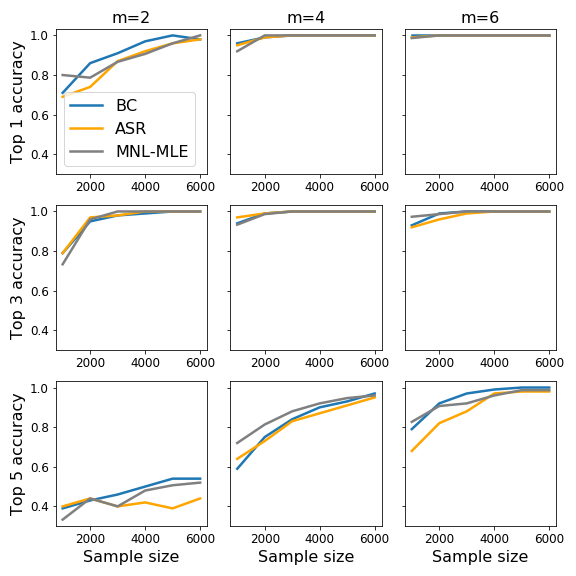}
\caption{\textbf{Irish West dataset ($n=9$):} %Exact top-$K$ accuracy against sample size. 
choice-based Borda count ({\color{blue}blue}) performs very similarly to MNL-MLE ({\color{gray}gray}) while the performance of ASR ({\color{orange}orange}) may diverge from MNL-MLE/choice-based Borda count.}
\label{fig:irish-west-experiment-compact}
\end{figure}

\textbf{Competitive accuracy:} Figure \ref{fig:irish-meath-experiment-compact} show the performance of the algorithms under the Irish-Meath dataset and Figure \ref{fig:irish-west-experiment-compact} shows the results for the Irish-West dataset. Our theoretical analysis in Section \ref{sect:connection-mle-borda} is reflected in our experimental findings: the performance of MNL-MLE and the Borda count algorithm are very similar given sufficiently large sample size. Spectral Ranking, on the other hand, may perform better or worse than MLE/Borda count depending on the dataset and the choice of $m$ and $K$. For many combinations of $m$ and $K$, we observe that the choice-based Borda count algorithm accurately recovers the top $K$ items and is highly competitive with MNL-MLE and Spectral Ranking. Notably, in most datasets, for smaller $K$ and large $m$, ithe Borda count algorithm has considerable advantages thanks to its accuracy and faster running time. 
In practice, this means choice-based Borda count is appropriate for applications where the statistician is interested in quickly determining a single (or a few) top candidate(s) from a large amount of data such as aggregating political surveys.

%% file: proofs/proofs.tex
\section{Proofs}\label{sect:proofs}
In this section, we will prove all the theoretical results of the paper. The first Section \ref{sect:IID-rum-properties}, however, details some of the important properties of IID-RUMs that are useful in proving the main theorems and lemmas. Many of these results can be of independent interest such as the \emph{Borda consistency} property which is shared by a very broad class of IID-RUMs. The reader may also skip to Section \ref{sect:error-bounds-top-K} for the proofs of the main results stated earlier in the paper.

\import{./}{iid_rum_closer_look}

\import{./}{proofs_upper_bound}

\import{./}{proofs_lower_bound}

\import{./}{proofs_menu_size}

\import{./}{proofs_connection}

%% file: proofs/iid_rum_closer_look.tex
\subsection{A closer look at IID-RUMs}
\label{sect:IID-rum-properties}
All IID-RUMs are in a family of choice models known as Fechnerian models \cite{becker1963stochastic}.
\begin{definition} A choice model $\rho$ is a Fechnerian model if there exists a function $F: \R \times \R \rightarrow [0,1]$ that is increasing in the first argument and decreasing in the second argument such that
$$ \rho(i|\{i, j\}) = F(U_i, U_j)\,. $$
\end{definition}
It is well known that Fechnerian models satisfy strongly stochastic transitivity (SST). That is, if $\rho(i|\{i, j\}) > \frac{1}{2}$ then $\rho(i|\{i, k\}) \geq \rho(j|\{j, k\})$ for all $k \neq i, j$. A weaker notion of stochastic transitivity is weak stochastic transitivity (WST): if $\rho(i|\{i, j\}) > \frac{1}{2}, \rho(j|\{j,k\}) > \frac{1}{2}$ then $\rho(i|\{i, k\}) > \frac{1}{2}$. The following lemma establishes that all IID-RUMs are Fechnerian models. 
\begin{lemma} All IID-RUMs are Fechnerian models.
\end{lemma}

\begin{proof} The proof is simply showing the existence of the function $F$ in the definition of Fechnerian models. Consider any two items $i$ and $j$
\begin{equation*}
\begin{aligned}
\rho(i|\{i, j\}) &= \PP(X_i > X_j) = \PP(U_i + \epsilon_i > U_j + \epsilon_j)\\
&= \PP(\epsilon_j - \epsilon_i < U_i - U_j)\\
&= F_{ij}(U_i - U_j)\,.
\end{aligned}
\end{equation*}
where $F_{ij}$ is the CDF of the random variable $\epsilon_j - \epsilon_i$. Note that all the $\epsilon_i$'s are identically distributed and thus $F = F_{ij}$ for all pairs $(i,j)$. This completes the proof.
\end{proof}

As the class of SST models are a special subclass of WST models, there exists a universal ordering among the items, induced by pairwise preference. Under IID-RUMs, this ordering the same as that of the partworth parameters. It's also notable that while IID-RUMs satisfy SST, RUMs with independent but not identically distributed noise (Independent RUMs) don't in general. In fact, Independent RUMs may not even satisfy WST. Next, we introduce a property of random utility models which we term the \textit{order preserving} property. 

\begin{definition} A choice model $\rho$ satisfies the order preserving property if
$$ \rho(i|\{i, j\}) > \frac{1}{2} \,\Rightarrow\, \rho(i|S) > \rho(j|S) \,\forall\, S: i, j \in S \,.$$
\end{definition}

The following lemma states that a broad class of IID-RUMs which include the MNL and Probit model satisfy the order preserving property.

\begin{lemma}\label{lem:iid-rum-order} All IID-RUMs whose noise distribution has absolutely continuous density function and support on the real line satisfy the order preserving property.
\end{lemma}

\begin{proof} Consider a pair of item $i, j$ such that $U_i > U_j$. The proof for the case when $S = \{i, j\}$ is trivial. We immediately have $\rho(i|\{i,j\}) > \frac{1}{2}$. Now consider any choice set $S$ containing both $i$ and $j$ where $|S| \geq 3$. Let $f_{\epsilon_i}$ denote the density function of the item-specific noise distribution of item $i$. Of course, under IID-RUM, $f_{\epsilon_i} = f_{\epsilon_j} = f$ for all $i\neq j$. We have
\begin{equation*}
    \begin{aligned}
    \rho(i | S) &= \PP(X_i > X_k , \forall k \in S\backslash \{i \})\\
    &= \PP(U_i + \epsilon_i > U_k + \epsilon_k, \forall k\in S\backslash\{i\})\\
    &= \PP(\epsilon_k < U_i - U_k + \epsilon_i, \forall k\in S\backslash\{i\})\\
    &= \int_\epsilon \PP(\epsilon_k < U_i - U_k + \epsilon_i, \forall k\in S\backslash\{i\}, \epsilon_i = \epsilon) d\epsilon\\
    &= \int_\epsilon \PP(\epsilon_k < U_i - U_k + \epsilon_i, \forall k\in S\backslash\{i\}| \epsilon_i = \epsilon) f_{\epsilon_i}(\epsilon) d\epsilon\\
    &= \int_\epsilon \bigg( \prod_{k\in S\backslash \{i\}} \PP(\epsilon_k < U_i - U_k  + \epsilon)  \bigg) f(\epsilon)d\epsilon \,.\\
    \end{aligned}
\end{equation*}
In the last equality, we have used the independence assumption of IID-RUMs. Further breaking down the product gives

\begin{equation}
\label{proof:order-preserving-eq1}
\begin{aligned}
&\int_\epsilon \bigg( \prod_{k\in S\backslash \{i\}} \PP(\epsilon_k < U_i - U_k  + \epsilon) \bigg) f(\epsilon)d\epsilon \\
&= \int_\epsilon \bigg( \prod_{k\in S\backslash \{i, j\}} \PP(\epsilon_k < U_i - U_k  + \epsilon)  \bigg) \PP(\epsilon_j < U_i - U_j + \epsilon) f(\epsilon)d\epsilon \,.\\
\end{aligned}
\end{equation}

By our assumption of identicallity and that the noise distribution has continuous density with support on the real line, $\PP(\epsilon_k < U_i - U_k + \epsilon) < \PP(\epsilon_k < U_j - U_k + \epsilon)$ for all $k\in S\backslash\{i, j\}$. 
On the other hand, for a random variable $\epsilon'$ distributed identically to the $\epsilon_i$'s,
$$ \PP(\epsilon_j < U_i - U_j + \epsilon) > \PP(\epsilon' < U_j - U_i + \epsilon) \quad \forall \epsilon\,. $$
Putting these two inequalities back into (\ref{proof:order-preserving-eq1}) gives
\begin{equation*}
\begin{aligned}
\rho(i | S) &= \int_\epsilon \bigg( \prod_{k\in S\backslash \{i, j\}} \PP(\epsilon_k < U_i - U_k  + \epsilon)  \bigg) \PP(\epsilon_j < U_i - U_j + \epsilon) f(\epsilon) d\epsilon\\
&> \int_\epsilon \bigg( \prod_{k\in S\backslash \{i, j\}} \PP(\epsilon_k < U_j - U_k  + \epsilon) \bigg) \PP(\epsilon' < U_j - U_i + \epsilon) f(\epsilon) d\epsilon\\
&= \rho(j | S)\,.
\end{aligned}
\end{equation*}
This completes the proof.\end{proof}

While the class of IID-RUMs whose noise distribution has continuous density function and support on the real line includes many commonly used IID-RUMs such as the MNL and Probit model, there are other IID-RUMs outside this class that also satisfy the order preserving property. An example is IID-Exponential RUM.

\begin{lemma}\label{lem:iid-exp-order-preserving} IID-Exponential RUM satisfies the order preserving property.
\end{lemma}

\begin{proof} Since the pairwise case ($m=2$) is trivial, we will focus on proving Lemma \ref{lem:iid-exp-order-preserving} for $m \geq 3$. Consider a menu $S\in \Cset^{(m)}: i, j \in S$, let $S' = S \backslash \{i, j\}$. For this proof, let us consider a `copy' of $i$ and call it $i'$. That is, $X_{i'}$ is distributed identically to $X_i$. We have
\begin{equation*}
\begin{aligned}
\rho(i | S) &= \Pr(X_i > \max\{ \{X_k\}_{k\in S'} \cup \{j\}\})\\
&= \int_{0}^{\infty} \Pr(X_k < U_i + \epsilon \,\forall k \in S', X_j < U_i + \epsilon) \cdot f(\epsilon) d\epsilon\\
&= \int_{0}^{\infty} \Pr(X_k < U_i + \epsilon \,\forall k\in S') \cdot \Pr(X_j < U_i + \epsilon) \cdot f(\epsilon)d\epsilon\\
&> \int_{0}^{\infty} \Pr(X_k < U_i + \epsilon \,\forall k\in S') \cdot \Pr(X_{i'} < U_i + \epsilon) \cdot f(\epsilon)d\epsilon\\
&> \int_{0}^{\infty} \Pr(X_k < U_j + \epsilon \,\forall k\in S') \cdot \Pr(X_{i'} < U_j + \epsilon) \cdot f(\epsilon)d\epsilon\\
&= \int_{0}^{\infty} \Pr(X_k < U_j + \epsilon \,\forall k \in S', X_{i'} < U_j + \epsilon) \cdot f(\epsilon)d\epsilon\\
&= \Pr(X_j > \max\{ \{X_k\}_{k\in S'} \cup \{i'\}\})\\
&= \rho(j|S)\,.
\end{aligned}
\end{equation*}
The first inequality holds because for any $\epsilon > U_j - U_i$, $\Pr(X_j < U_i + \epsilon) > \Pr(X_{i'} < U_i + \epsilon)$. The second inequality holds because for any $\epsilon > \max\{0,\max\{U_k - U_j\}_{k\in S' \cup\{j\}}\}$, $\Pr(X_k < U_i + \epsilon) > \Pr(X_k < U_j + \epsilon)$ for all $k \in S'$ and $\Pr(X_{i'} < U_j + \epsilon) > \Pr(X_{i'} < U_j + \epsilon)$. This completes the proof.
\end{proof}

Recall the Borda Consistency property introduced in Section \ref{subsect:generalized-borda-score}. One can show that a broad class of IID-RUMs which contains many commonly used RUMs such as the MNL and the Probit model satisfies Borda Consistency, the key property that ensures correctness of Generalized Borda Count.

\begin{theorem} Any IID-RUM $\rho$ whose noise distribution has continuous density function and support on the real line satisfies Borda consistency.
\end{theorem}

\begin{proof} Note that $|\Cset_i^{(m)}| =  |\Cset_j^{(j)}| = {n-1\choose m-1}$ for any $i, j$ and $m$. The proof for $m = 2$ is trivial so we will focus on $m \geq 3$. By definition,
\begin{equation*}
\begin{aligned}
\tau^{(m)}_i &= \frac{1}{|\Cset_i^{(m)}|} \sum_{S \in \Cset_i^{(m)}} \rho(i | S)\\
&= \frac{1}{|\Cset_i^{(m)}|} \bigg( \sum_{S: |S|=m, i, j \in S }\rho(i | S) + \sum_{S: |S|=m, i\in S, j\notin S} \rho(i|S) \bigg)\,.
\end{aligned}
\end{equation*}
By the ordering preserving property of IID-RUMs, for any $S: i, j \in S$, $p(\rho|S) > \rho(j|S)$. To prove that $\tau^{(m)}_i > \tau^{(m)}_j$, we only need to prove that for any $S': |S'| = m-1; i, j \notin S'$,
$$ \rho(i|S' \cup \{i\}) > \rho(j|S'\cup \{j\})\,. $$
Let $f_{\epsilon}$ denote the density function of the noise distribution, we have
\begin{equation*}
\begin{aligned}
\rho(i | S'\cup \{i\}) &= \PP(X_i > X_k \forall k \in S')\\
&= \PP(U_i + \epsilon_i > U_k + \epsilon_k \forall k \in S')\\
&= \PP(\epsilon_k < U_i - U_k + \epsilon_i \forall k \in S')\\
&= \int_{-\infty}^{\infty} \PP(\epsilon_k < U_i - U_k + \epsilon_i, \epsilon_i = \epsilon)\\
&= \int_{-\infty}^{\infty} \PP(\epsilon_k < U_i - U_k + \epsilon) f(\epsilon) d\epsilon\\
&> \int_{-\infty}^{\infty} \PP(\epsilon_k < U_j - U_k + \epsilon) f(\epsilon) d\epsilon\\
&= \rho(j|S'\cup \{j\}) \,. 
\end{aligned}
\end{equation*}
This completes the proof.\end{proof}

We also have a more general theorem that characterizes the class of IID-RUMs, beyond those whose noise distribution has continuous density function and support on the real line, that satisfy Borda consistency
\begin{theorem}\label{thm:consistency-borda-score-general} Consider an IID-RUM $\rho$. Fix a menu size $m$. If, for any pair of item $i, j$ where $U_i > U_j$,
\begin{itemize} 
    \item There exists a menu $S \in \Cset^{(m)}: i, j\in S$ such that $\rho(i|S) \neq \rho(j|S)$, or
    \item $m \geq 3$ and there exists a menu $S'\in \Cset^{(m-1)}: i, j \notin S$ such that $\rho(i|S\cup\{i\}) \neq \rho(j|S\cup\{j\})$,
\end{itemize}
then $\rho$ satisfies Borda consistency.
\end{theorem}

\begin{proof} Consider an IID-RUM $\rho$ and a fixed menu size $m$. For any pair of items $i, j$ where $U_i > U_j$, one can easily check that
$$ \rho(i|S) \geq \rho(j|S) \,\forall S \in \Cset^{(m)}: i, j\in S $$
and
$$ \rho(i|S'\cup\{i\}) \geq \rho(j|S'\cup \{j\}) \,\forall S'\in \Cset^{(m-1)}: i, j\notin S'\,.$$
If there exists a menu $S\in\Cset^{(m)}:i, j\in S$ where $\rho(i|S) \neq \rho(j|S)$, then $\rho(i|S) > \rho(j|S)$. Similarly, if there exists a menu $S'\in\Cset^{(m-1)}:i, j\notin S$ where $\rho(i|S\cup \{i\}) \neq \rho(j|S\cup\{j\})$ then $\rho(i|S\cup\{i\}) > \rho(j|S\cup\{j\})$. If either of these cases hold, we have
\begin{equation*}
\begin{aligned}
\tau^{(m)}_i &= \frac{1}{|\Cset_i^{(m)}|} \sum_{S \in \Cset_i^{(m)}} \rho(i | S)\\
&= \frac{1}{|\Cset_i^{(m)}|} \bigg( \sum_{S: |S|=m, i, j \in S }\rho(i | S) + \sum_{S: |S|=m, i\in S, j\notin S} \rho(i|S) \bigg)\\
&> \frac{1}{|\Cset_j^{(m)}|} \bigg( \sum_{S: |S|=m, i, j \in S }\rho(j | S) + \sum_{S: |S|=m, j\in S, i\notin S} \rho(j|S) \bigg) = \tau^{(m)}_j\,.
\end{aligned}
\end{equation*}
This completes the proof.\end{proof}

%% file: proofs/proofs_upper_bound.tex
\subsection{Sample complexity for exact top-$K$ recovery}
\label{sect:error-bounds-top-K}
In this section we prove all theorems stated in Section \ref{sect:upper-bound} as well as the sample complexity for approximate top-$K$ ranking of the choice based Borda count algorithm.

\begin{replemma}{thm:error-bound-gen-borda}
Consider an IID-RUM that satisfies Borda consistency per Definition \ref{def:borda-consistency}. Assume input choice data with menu size $m$ is generated according to the sampling model described in Section \ref{subsect:sampling-model}. For any two items $i$ and $j$ where $\tau_i^{(m)} > \tau_j^{(m)}$, the choice-based Borda count algorithm satisfies
$$\PP(\hat W_j > \hat W_i) \leq \exp{\frac{-3pR{n\choose m} {m(\tau_i^{(m)}-\tau_j^{(m)})}^2 }{8n(\tau_i^{(m)} +\tau_j^{(m)})}} \,.$$
\end{replemma}

\begin{proof} 
Let $\calE$ denote the event $\{i \in \Ss^*_K, j\notin \Ss^*_K\}$. We wish to prove that if $\tau_i - \tau_j > {\Delta_K^{(m)}}$, then with sufficiently large $p$ and $R$, $\hat W_i$ is smaller than $\hat W_j$ with very small probability. The main external concentration inequality used in this proof is Bernstein's inequality (cf. Theorem 2.8.4 \cite{vershynin2018high}). We start our proof by expanding on the probability of misranking $i$ and $j$.
\begin{equation}\label{eqn:bound-easy-expand}
\begin{aligned}
&\PP\bigg(\hat W_j > \hat W_i \,\lvert\, \calE\bigg)\\
&= \PP\bigg(\sum_{r=1}^R \sum_{S\in \Cset_j^{(m)}} \underbrace{\mathbf{1}[y_S^{(r)} = j]}_{X_S^{j,r}} - \sum_{r=1}^R \sum_{S\in \Cset_i^{(m)}} \underbrace{\mathbf{1}[y_S^{(r)} = i]}_{X_S^{i,r}} > 0 \,|\, \calE   \bigg)\\
&= \PP\bigg(\sum_{r=1}^R\sum_{S\in\Cset_j^{(m)}} X_S^{j,r} - \E[X_S^{j,r}] + \E[X_S^{j,r}] - \sum_{r=1}^{R}\sum_{S\in\Cset_i^{(m)}} X_S^{i,r} - \E[X_S^{i,r}] + \E[X_S^{i,r}] > 0 \,|\,\calE \bigg)\\
&= \PP\bigg(\sum_{r=1}^R\sum_{S\in\Cset_j^{(m)}} X_S^{j,r} - \E[X_S^{j,r}] - \sum_{r=1}^{R}\sum_{S\in\Cset_i^{(m)}} X_S^{i,r} - \E[X_S^{i,r}]> \sum_{r=1}^R\sum_{S\in\Cset_i^{(m)}} \E[X_S^{i,r}]  - \sum_{r=1}^{R}\sum_{S\in\Cset_j^{(m)}}  \E[X_S^{j,r}] \,|\,\calE \bigg)\\
&= \PP\bigg(\sum_{r=1}^R\sum_{S\in\Cset_j^{(m)}} X_S^{j,r} - \E[X_S^{j,r}] - \sum_{r=1}^{R}\sum_{S\in\Cset_i^{(m)}} X_S^{i,r} - \E[X_S^{i,r}] > pR{n-1\choose m-1}(\tau^{(m)}_i - \tau^{(m)}_j) \bigg)\,.
\end{aligned}
\end{equation}
In the last equality, we have used the fact that $\sum_{r=1}^R\sum_{S\in\Cset_i^{(m)}}\E[X_S^{i,r}] - \sum_{r=1}^R\sum_{S\in\Cset_j^{(m)}}\E[X_S^{j,r}] = pR{n-1 \choose m-1}(\tau_i^{(m)} - \tau_j^{(m)}) \geq pR{n-1\choose m-1} (\tau^{(m)}_i - \tau^{(m)}_j)$. Now, we will expand the two terms on the LHS of the inequality as
\begin{equation}
\label{eqn:to-apply-bernstein}
\begin{aligned}
&\PP\bigg(\hat W_j > \hat W_j \,\lvert\, \calE\bigg)\\
&= \PP\bigg( \sum_{S\in \Cset_j^{(m)}: i\notin S} \sum_{r=1}^R \big[X_S^{j,r} - \E[X_S^{j,r}] \big] - \sum_{S\in \Cset_i^{(m)}: j\notin S} \sum_{r=1}^R \big[X_S^{i,r} - \E[X_S^{i,r}] \big]\\
& + \sum_{S \in \Cset^{(m)}: i, j\in S} \sum_{r=1}^R  \big[(X_S^{j,r} - X_S^{i,r}) - (\E[X_S^{j,r} -X_S^{i,r}]) \big] > pR{n-1\choose m-1}(\tau^{(m)}_i - \tau^{(m)}_j)  \bigg)\,.
\end{aligned}
\end{equation}

Note that with the above decomposition, the terms $\{X_S^{j,r}\}_{S\in \Cset^{(m)}_j: i\notin S}, \{X_S^{i,r}\}_{S\in \Cset^{(m)}_i: j\notin S}, \{X_S^{j,r}-X_S^{i,r}\}_{S\in \Cset^{(m)}: i, j \in S}$ are all mutually independent. To apply Bernstein's inequality, we need to evaluate (or upper bound) the following.
$$\sum_{S\in \Cset_j^{(m)}: i\notin S} \sum_{r=1}^R \E[(X_S^{j,r})^2]+\sum_{S\in \Cset_i^{(m)}: j\notin S} \sum_{r=1}^R \E[(X_S^{i,r})^2] +\sum_{S \in \Cset^{(m)}: i, j\in S} \sum_{r=1}^R  \E[(X_S^{j,r} - X_S^{i,r})^2] \,. $$
We can easily see that for $S \in \Cset^{(m)}_j: i\notin S$,
$$ X_S^{j, r} = \begin{cases} 1 &\text{ with probability } p\cdot \rho(j|S)\\
    0 &\text{ with probability } 1 - p\cdot \rho(j|S)\\
\end{cases} $$
and for $S\in \Cset^{(m)}: i, j \in S$,
$$ X_S^{j, r} - X_S^{i,r} = \begin{cases} 1 &\text{ with probability } p\cdot \rho(j|S)\\
    -1 &\text{ with probability } p\cdot \rho(i|S)\\
    0 &\text{ with probability } 1 - p\cdot \rho(j|S) -p\cdot \rho(i|S)\\
\end{cases}\,. $$
As such $\E[(X_S^{j, r})^2] =  p\cdot\rho(j|S)$ and $\E[(X_S^{j,r} - X_S^{i,r})^2] = p\cdot(\rho(j|S) + \rho(i|S)) $. We have
\begin{equation*}
\begin{aligned}
&\sum_{S\in \Cset_j^{(m)}: i\notin S} \sum_{r=1}^R \E[(X_S^{j,r})^2]+\sum_{S\in \Cset_i^{(m)}: j\notin S} \sum_{r=1}^R \E[(X_S^{i, r})^2] +\sum_{S \in \Cset^{(m)}: i, j\in S} \sum_{r=1}^R  \E[(X_S^{j,r} - X_S^{i,r})^2] \\
&= \sum_{S\in \Cset_j^{(m)}: i\notin S} \sum_{r=1}^R p\cdot\rho(j|S) + \sum_{S\in \Cset_i^{(m)}: j\notin S} \sum_{r=1}^R p\cdot\rho(i|S) + \sum_{S\in \Cset^{(m)}: i, j \in S} \sum_{r=1}^R p\cdot(\rho(j|S) + \rho(i|S))\\
&= \sum_{S\in \Cset_j^{(m)}} \sum_{r=1}^R p\cdot\rho(j|S) + \sum_{S\in \Cset_i^{(m)}} \sum_{r=1}^R p\cdot\rho(i|S)\\
&= pR{n-1\choose m-1}\tau^{(m)}_i  + pR{n-1\choose m-1}\tau^{(m)}_j\\
&= pR{n-1\choose m-1}(\tau^{(m)}_i + \tau^{(m)}_j) \,. \\
\end{aligned}
\end{equation*}
Now, applying Bernstein's inequality to (\ref{eqn:to-apply-bernstein}) directly yields:
\begin{equation*}
\begin{aligned}
&\PP\bigg(\hat W_j > \hat W_j \,\lvert\, \calE\bigg)\\
&\leq \exp{-\frac{\big(pR{n-1\choose m-1}{(\tau^{(m)}_i - \tau^{(m)}_j)}\big)^2}{2(pR{n-1\choose m-1}(\tau^{(m)}_i + \tau^{(m)}_j)+ \frac{pR{n-1\choose m-1}{(\tau^{(m)}_i - \tau^{(m)}_j) }}{3} )}}\\
&= \exp{-\frac{pR{n-1\choose m-1}{(\tau^{(m)}_i - \tau^{(m)}_j)}^2}{2(\tau^{(m)}_i + \tau^{(m)}_j)+\frac{2 (\tau^{(m)}_i - \tau^{(m)}_j) }{3}}}\\
&\leq \exp{-\frac{pR{n-1\choose m-1}(\tau^{(m)}_i - \tau^{(m)}_j)^2}{8/3(\tau_i^{(m)} +\tau_j^{(m)} ) }}\,.
\end{aligned}
\end{equation*}
This completes the proof. \end{proof}

Next we present the proof of Lemma \ref{lem:simple-inequality}.
\begin{replemma}{lem:simple-inequality}
Consider an IID-RUM that satisfies Borda consistency per definition (\ref{def:borda-consistency}). Fix a $K$, we have
$$ \frac{\tau^{(m)}_K + \tau^{(m)}_{K+1}}{{\Delta_K^{(m)}}^2 } = \max_{i\in \Ss^*_K, j\notin \Ss^*_K} \bigg\{ \frac{\tau^{(m)}_i + \tau^{(m)}_j}{(\tau^{(m)}_i - \tau^{(m)}_j)^2 } \bigg\} \, .$$
\end{replemma}

% \begin{replemma}\label{lem:simple-inequality}
% Consider an IID-RUM $\rho$ that satisfies Borda consistency. Consider two items $i\in \Ss_K^*$ and $j\notin \Ss_K^*$. Given choice data with menu size $m$ generated according to the sampling model described in section (\ref{subsect:sampling-model}), the Borda Count algorithm satisfies:
% $$ \PP(\hat W_j > \hat W_i \,\lvert\, i \in \Ss^*_K, j\notin \Ss^*_K) \leq \exp{\frac{-3pR{n-1\choose m-1} (\tau^{(m)}_i - \tau^{(m)}_j)^2 }{8(\tau_i^{(m)} +\tau_j^{(m)} )}}$$
% \end{lemma}

\begin{proof}
Firstly, it is easy to see that, fixing an $i\in \Ss_K^*$, for all $j\notin \Ss_K^*$,
$$ \frac{\tau_i^{(m)} + \tau_{K+1}^{(m)} }{(\tau_i^{(m)} - \tau_{K+1}^{(m)})^2 } \geq \frac{\tau_i^{(m)} + \tau_j^{(m)} }{(\tau_i^{(m)} - \tau_j^{(m)})^2} \,.$$

It remains to prove that for any $i \in \Ss_K^*$,
$$ \frac{\tau_K^{(m)} + \tau_{K+1}^{(m)}}{(\tau_K^{(m)} - \tau_{K+1}^{(m)})^2} \geq \frac{\tau_i^{(m)} + \tau_{K+1}^{(m)} }{(\tau_i^{(m)} - \tau_{K+1}^{(m)})^2 }\,. $$
To declutter the notation, we'll remove the superscript $m$. Let $\Delta_{iK} = \tau_i - \tau_K$ for some $i \in \Ss_K^*$. For any $i \in \Ss_K^* \backslash \{K\}$, we have
\begin{equation*}
\begin{aligned}
\frac{\tau_K + \tau_{K+1}}{\Delta_K^2} &\geq \frac{\tau_i + \tau_{K+1}}{(\tau_i - \tau_{K+1})^2}\\
\Leftrightarrow \frac{\tau_K + \tau_{K+1}}{\Delta_K^2} &\geq \frac{\tau_K + \Delta_{iK} + \tau_{K+1}}{(\Delta_{iK} + \Delta_K)^2}\\
\Leftrightarrow \frac{\tau_K + \tau_{K+1}}{\Delta_K^2} &\geq \frac{\tau_K + \Delta_{iK} + \tau_{K+1}}{\Delta^2_{iK} + \Delta^2_K + 2\Delta_{iK}\Delta_K}\\
\Leftrightarrow \tau_K\Delta_{iK}^2 + \tau_K\Delta_K^2 +2\tau_K\Delta_{iK}\Delta_K +\tau_{K+1}\Delta_{iK}^2 &+\\
\tau_{K+1}\Delta_K^2 +2\tau_{K+1}\Delta_{iK}\Delta_K &\geq \tau_K\Delta_K^2 + \tau_{K+1}\Delta_K^2+\Delta_{iK}\Delta_K^2 \\
\Leftrightarrow \tau_K\Delta_{iK}^2 +2\tau_K\Delta_{iK}\Delta_K +\tau_{K+1}\Delta_{iK}^2 +2\tau_{K+1}\Delta_{iK}\Delta_K &\geq \Delta_{iK}\Delta_K^2\,.
\end{aligned}
\end{equation*}
In deriving the last statement, the first two terms on the RHS get canceled out. Note that $2\tau_K\Delta_{iK}\Delta_K \geq 2\Delta_K\Delta_{iK}\Delta_K \geq \Delta_{iK}\Delta_K^2$. This completes the proof.
\end{proof}

\begin{reptheorem}{cor:sample-complexity-gbs}
Assume the conditions of Lemma \ref{thm:error-bound-gen-borda}. Given sufficiently large $p, R$ such that $pR{n \choose m} \geq \frac{8n \log n}{m{\Delta_K^{(m)}}^2}\cdot(\Delta_K^{(m)} +2\tau_{K+1}^{(m)})$, the choice-based Borda count algorithm correctly identifies all of the top $K$ items with probability at least $1-O(\frac{K}{n^2})$.
\end{reptheorem}

\begin{proof} The exponential bound in Lemma \ref{thm:error-bound-gen-borda} holds simultaneously for all pairs $i \in \Ss_K^*, j\notin \Ss_K^*$ given sufficiently large $p,R$ such that
$$ pR{n-1\choose m-1} \geq \max_{i\in \Ss^*_K, j\notin \Ss^*_K} \frac{8\log n}{(\tau^{(m)}_i - \tau^{(m)}_j)^2} \cdot (\tau_i^{(m)} +\tau_j^{(m)}) \,.$$
From Lemma (\ref{lem:simple-inequality}), we have
$$ \max_{i\in \Ss^*_K, j\notin \Ss^*_K} \frac{\tau_i^{(m)} + \tau_j^{(m)} }{(\tau_i^{(m)} - \tau_j^{(m)})^2 } = \frac{\tau_K^{(m)} +\tau_{K+1}^{(m)}}{{\Delta_K^{(m)}}^2} \,.$$
This means that if
$$ pR{n-1\choose m-1} \geq 8\log n \cdot\frac{\tau_K^{(m)}+\tau_{K+1}^{(m)} }{{\Delta_K^{(m)}}^2 } $$
then
$$ \Pr(\hat W_i < \hat W_j) \leq \frac{1}{n^3}  \,.$$
Apply union bound over all pairs $i, j$ such that $i \in \Ss_K^*, j\notin \Ss_K^*$. Then the event
$$ \hat W_i > \hat W_j \,\,\forall i\in \Ss_{K}^*, j\notin \Ss_{K}^* $$
happens with probability at least $1-O(\frac{K}{n^2})$. 
\end{proof}

\subsection{Sample complexity for approximate top-$K$ recovery}

In practice, some error in the top $K$ estimate could be tolerable. This is known as approximate top-$K$ ranking. The metric of interest in approximate top-$K$ ranking is the edit distance between the estimate $\hat\Ss_K$ and the true top $K$ items $\Ss^*_K$:
$$ D_{01}(\hat \Ss_K, \Ss_K^*) = K - \lvert \hat\Ss_K \cup \Ss_K \rvert \,. $$
One can see that correctly separating the top $K-h$ items and the bottom $n-K-h$ items, i.e., unable to identify at most $h$ of the top $K$ items, is sufficient to guarantee that the approximate loss is bounded by $h$.
Considering this, the fundamental quantity that determines the hardness of approximate top-$K$ ranking is the gap between the generalized Borda Score of the $K-h$ item and that of the $K+h+1$ item. For convenience, let us denote such quantity as 
$$\Delta^{(m)}_{K,h} =  \tau^{(m)}_{K-h-1} - \tau^{(m)}_{K+h+1}\,. $$
Clearly, this quantity generalizes $\Delta_K^{(m)}$ as $\Delta_{K,0}^{(m)} =\Delta_K^{(m)}$. We are interested in bounding the probability that, given an error threshold $h$, Generalized Borda Count fails to output an estimate with loss bounded by $h$. Namely,
$$ \Pr(D_{01}(\hat \Ss_K, \Ss_K^*) \leq h)\,. $$
Building on Lemma \ref{thm:error-bound-gen-borda}, we obtain the following sample complexity of Generalized Borda Count for approximate top-$K$ ranking.

\begin{theorem}\label{cor:sample-complexity-borda-approximate}
Assume the conditions of Lemma \ref{thm:error-bound-gen-borda}. Fix an error threshold $h$. Given sufficiently large $p, R$ such that $pR{n \choose m} \geq \frac{8n \log n}{m{\Delta_{K,h}^{(m)}}}\cdot(1+\frac{\tau_{K+h+1}^{(m)}}{\Delta_{K,h}^{(m)} } )$, the choice based Borda count algorithm outputs a set $\hat\Ss_K$ that satisfies
$$ D_H(\Ss^*_K, \hat\Ss_K) \leq h $$
with probability at least $1-O(\frac{K}{n^2})$.
\end{theorem}

\begin{proof} Applying Lemma \ref{thm:error-bound-gen-borda}, we have for every pair $i, j$ where $i \in \Ss_{K-h}^*$, $j\notin \Ss_{K+h}^*$,
$$ \Pr(\hat W_j > \hat W_i) \leq \exp{-\frac{3pR{n-1\choose m-1}(\tau^{(m)}_{i}-\tau_j^{(m)})^2 }{8(\tau^{(m)}_{i}+\tau_j^{(m)}) }   }\,.$$

Following the same argument as in the proof of Theorem \ref{cor:sample-complexity-gbs}, one can show that
$$ \frac{\tau_{K-h}+ \tau_{K+h+1}}{\Delta^2_{K,h}} = \max_{i\in \Ss_{K-h}^*, j\notin \Ss_{K+h}^*}\frac{\tau_i^{(m)} +\tau_j^{(m)}}{(\tau^{(m)}_{i}-\tau_j^{(m)})^2 }\,. $$
The rest of the proof follows by showing that given the condition on $p, R$ in the theorem statement, the probability that Generalized Borda Count wrongly flips the relative order of any two items $i \in \Ss_{K-h}^*, j \notin \Ss_{K+h}^*$ is upper bounded by $O(\frac{1}{n^3})$. Applying union bound over all such pairs of $i, j$ completes the proof. 
\end{proof}

In short, Generalized Borda Count has the following sample complexity for approximate top-$K$ ranking
$$ O\bigg( \frac{n\log n}{m\Delta_{K,h}^{(m)}} \cdot (1+\frac{\tau_{K+h+1}^{(m)}}{\Delta_{K,h}^{(m)}} \bigg)\,. $$

%% file: proofs/proofs_lower_bound.tex
\subsection{Lower bound on the sample complexity of top-$K$ recovery}

\subsubsection{Preliminaries and notations}
We first restate a version of Fano's inequality \cite{cover1999elements} which will be useful to the construction of our lower bound:

\begin{lemma} [Fano's inequality] Consider a set of $L$ distributions $\{\Pr^1,\ldots, \Pr^L\}$. Suppose that we observe a random variable (or a set of random variables) $Y$ that was generated by first picking an index $A \in \{1,\ldots,L\}$ uniformly at random and then $Y \sim \Pr^A$. Fano's inequality states that any hypothesis test $\phi$ for this problem has an error probability lower bounded as
$$ \Pr[\phi(Y) \neq A] \geq 1 - \frac{\max_{a,b\in [L], a \neq b} \kl(\Pr^a(Y)\rVert \Pr^b(Y)) + \log 2 }{\log L}\,. $$
\end{lemma}

To obtain a lower bound in the form of Theorem \ref{thm:lower-bound-all}, we need to construct an IID-RUM such that any method requires $\Omega(n\log n)$ examples in order to accurately recover the top $K$ items. For this purpose, we look for a model within the Multinomial Logit family. It is well known that the MNL model is an instance of IID-RUM where the noise distribution is the standard Gumbel distribution. While the MNL model has a random utility characterization, it is often more convenient to describe the MNL model in terms of weighted probabilities. Specifically, an MNL model over $n$ items can be parameterized by a set of $n$ positive real numbers $w_1, \ldots, w_n$ called weights. There is also a well defined relation between the weighted probabilities representation and the utility partworth representations. Specifically, the resulting choice probability is defined to be:
$$\rho(i|S) = \frac{w_i}{\sum_{k\in S} w_k} = \frac{e^{U_i}}{\sum_{k\in S}e^{U_k} } $$
Now consider a special sub-class of MNL models that is defined by 3 parameters $(v, \delta, \Ss^*_K)$ for some $0 < \delta < v$ and a set $\Ss^*_K$ of size $K$. The item-specific weights are then defined as follows.
$$ w_i = \begin{cases}v + \delta &\text{if } i \in \Ss^*_K\\
v &\text{otherwise}
\end{cases} \,.$$
That is, all the top items have the same weights and all the bottom items have the same (but smaller) weight. For any $A \in [K, \ldots, n]$ let $\mathcal{M}^A$ be a special MNL model parametrized by $(v, \delta, \{1,\ldots, K-1, A\})$ as described above. One can see that Fano's inequality can be used to lower bound the error probability of any hypothesis test for the identity of $A$. That is, to lower bound the probability that any estimator fails to correctly identify the $K$-th item.

Recall our uniform sampling model described in Section \ref{subsect:sampling-model}, let us use $\{Y^r_S\}_{S\in \Cset^{(m)}, r=1:R}$ to denote the observed choice data where:
$$ Y_S^r = \begin{cases} y &\text{for some } y \in S \text{ if menu $S$ is offered in round }r\\
    0 &\text{if menu $S$ is not offered in round }r
\end{cases}\,. $$
As a short hand, denote $\rho^a$ as the choice rule corresponding to $\calM^a$ and $\rho^b$ corresponding to $\calM^b$. Let $\Pr^a(\{Y_S^r\})$ denote the likelihood of choice data $\{Y_S^r\}$ under choice model $\calM^a$ (define analogously for $\calM^b$).

\subsubsection{Exact top-$K$ recovery}\label{subsub:proof-exact-top-k}
The following pair of lemma and theorem establish the lower bound for exact top-$K$ recovery via showing an upper bound on the KL divergence between any two models $\calM^a, \calM^b$ and then applying Fano's inequality.
    
\begin{lemma}\label{lem:kl-div}
    Assuming the same sampling model as in Theorem \ref{thm:error-bound-gen-borda}, we have, for any $a, b \in [K, \ldots, n]$,
    $$ \kl(\Pr^a(\{Y_S^r\})\rVert \Pr^b(\{Y_S^r\})) \leq 4pR{n-1\choose m-1}\frac{{\Delta_K^{(m)}}^2}{\tau_{K}^{(m)}} \,. $$
\end{lemma}

\begin{reptheorem}{thm:lower-bound-all}
Consider the sampling model described in Section \ref{subsect:sampling-model}.
There exists a choice model $\calM^*$ within the class of Multinomial Logit Models (MNLs) such that for $n \geq 20$, if $pR{n\choose m} \leq \frac{n\log n }{8}\cdot\frac{\tau_{K}^{(m)}}{m{\Delta_K^{(m)}}^2}$ then any statistical estimator fails to correctly identify all of the top $K$ items with probability at least $\frac{1}{12}$.
\end{reptheorem}

Once Lemma \ref{lem:kl-div} has been established, we can prove Theorem \ref{thm:lower-bound-all} as follows. For any $n \geq 20$, if ${\Delta_K^{(m)}}^2 \leq \frac{\tau_{K}^{(m)}\cdot \log n}{8 pR{n-1\choose m-1}}$, Fano's inequality gives us a lower bound on the error of any estimator for finding the top $K$ items under $\calM(v, \delta, \Ss^*_K)$:
$$ \Pr(\hat \Ss_K \neq \Ss^*_K) \geq 1- \frac{0.5\log n +\log 2}{\log (n-K+1)} \geq 1- \frac{0.5\log n +\log 2}{\log (n/2+1)} \geq 0.086 \,. $$
The rest of the proof of Theorem \ref{thm:lower-bound-all} involves simple algebra, noting that ${n\choose m} = {n-1\choose m-1}\cdot \frac{n}{m}$. We now move on to the proof of Lemma \ref{lem:kl-div}.

\begin{proof} (Proof of Lemma \ref{lem:kl-div}). Let $\hat\C^{(m,r)}$ denote the set of menus of size $m$ being picked in round $r$ under the sampling model described in Section \ref{subsect:sampling-model}. Let us first decompose $\kl(\Pr^a(\{Y_S^r\})\rVert\Pr^b(\{Y_S^r\}))$ into a more manageable form.
\begin{equation*}
    \begin{aligned}
        &\kl(\Pr^a(\{Y_S^r\})\rVert\Pr^b(\{Y_S^r\}))\\
        &[\text{thanks to mutual independence among } \{Y_S^r\}]\\
        &=\sum_{r=1}^R \sum_{S\in\Cset^{(m)}} \kl(\Pr^a(Y_S^r) \rVert \Pr^b(Y_S^r))\\
        &=\sum_{r=1}^R \sum_{S\in\Cset^{(m)}} \kl(\Pr^a(Y_S^r|S\in \Cset_r^{(m,r)}) \Pr^a(S\in\hat\Cset^{(m)}) \rVert \Pr^b(Y_S^r|S\in \hat\Cset^{(m,r)}) \Pr^a(S\in\hat\Cset^{(m)}))\\
        &=\sum_{r=1}^R \sum_{S\in\Cset^{(m)}} \kl(\Pr^a(Y_S^r|S\in \hat\Cset^{(m,r)}) \Pr^a(S\in\hat\Cset^{(m,r)}) \rVert \Pr^b(Y_S^r|S\in \hat\Cset_r^{(m,r)}) \Pr^b(S\in\hat\Cset^{(m,r)}))\\
        &[\text{Using the fact that } \Pr^a(S\in\hat\Cset^{(m,r)}) = \Pr^b(S\in \hat\Cset^{(m,r)}) = p]\\
        &= pR\sum_{S\in \Cset^{(m)}}\kl(\Pr^a(Y_S^1|S\in\hat\Cset^{(m,1)})\rVert \Pr^b(Y_S^1|S\in\hat\Cset^{(m,1)}))\,.
    \end{aligned}
\end{equation*}
To handle the above sum, partition $\Cset^{(m)}$ into 4 subsets $\{S \in \Cset^{(m)}: a, b\notin S \}, \{S \in \Cset^{(m)}: a\in S, b\notin S \}, \{S \in \Cset^{(m)}: a\notin S, b \in S \}, \{S \in \Cset^{(m)}: a, b\in S \}$. It is easy to see that for $S : a, b\notin S$, the corresponding KL divergence terms are all 0. Hence, we only have to work with other 3 subsets. Before going into the details, let us define some shorthand notations: let $w^a(S)$ denote the sum of the weights of the items in menu $S$ under $\calM^a$ and $S + a$ denote set union. 

\textbf{Subset 1:} $\{S\in\Cset^{(m)}: a, b\in S \}$
\begin{equation*}
    \begin{aligned}
        &\sum_{S\in \Cset^{(m)}: a, b\in S}\kl(\Pr^a(Y_S^1|S\in\hat\Cset^{(m,1)})\rVert \Pr^b(Y_S^1|S\in\hat\Cset^{(m,1)}))\\
        &[\text{Note that $\rho^a(x|S) = \rho^b(x|S) \quad\forall x\neq a, b$ and $ w^a(S) = w^b(S)$ for all $S: a,b\in S$}]\\
        &=\sum_{S\in \Cset^{(m)}: a, b \in S} \rho^a(a|S)\log\frac{\rho^a(a|S)}{\rho^b(a|S)} + \rho^a(b|S)\log\frac{\rho^a(b|S)}{\rho^b(b|S)}\\
        &=\sum_{S\in \Cset^{(m)}: a, b \in S} \frac{v+\delta}{w^a(S)}\log\frac{v+\delta}{v} + \frac{v}{w^a(S)}\log\frac{v}{v+\delta}\\
        &= \sum_{S\in \Cset^{(m)}: a, b \in S} \frac{\delta}{w^a(S)}\log\frac{v+\delta}{v}\,.
    \end{aligned}
\end{equation*}

\textbf{Subset 2:} $\{S\in\Cset^{(m)}: a \in S, b\notin S \}$. Firstly, it is easy to verify the following useful identities: $w^a(S' + a) = w^b(S' + b), w^a(S'+b) = w^b(S'+a) \forall S': a, b\notin S'$. Furthermore, $w^a(x) = w^b(x) \forall x\neq a, b$. Now we have
\begin{equation*}
    \begin{aligned}
        &\sum_{S\in \Cset^{(m)}: a\in S, b\notin S}\kl(\Pr^a(Y_S^1|S\in\hat\Cset^{(m,1)})\rVert \Pr^b(Y_S^1|S\in\hat\Cset^{(m,1)}))\\
        &=\sum_{S'\in \Cset^{(m-1)}: a, b\notin S'} \bigg[ \bigg(\sum_{x\in S'} \rho^a(x|S'+a) \log\frac{\rho^a(x|S'+a)}{\rho^b(x|S'+a)}\bigg) + \rho^a(a|S'+a)\log \frac{\rho^a(a|S'+a)}{\rho^b(a|S'+a)} \bigg] \\
        &=\sum_{S'\in \Cset^{(m-1)}: a, b\notin S'} \bigg[\bigg(\sum_{x\in S'} \frac{w^a(x)}{w^a(S'+a)} \log\frac{w^b(S'+a)}{w^a(S'+a)} \bigg)+ \frac{v+\delta}{w^a(S'+a)}\log \frac{v+\delta}{v}\cdot\frac{w^b(S'+a)}{w^a(S'+a)} \bigg] \\
        &=\sum_{S'\in \Cset^{(m-1)}: a, b\notin S'} \bigg[\bigg( \sum_{x\in S'} \frac{w^a(x)}{w^a(S'+a)} \log\frac{w^a(S'+b)}{w^a(S'+a)}\bigg) + \frac{v+\delta}{w^a(S'+a)}\log \frac{v+\delta}{v}\cdot\frac{w^a(S'+b)}{w^a(S'+a)} \bigg] \,.
    \end{aligned}
\end{equation*}

\textbf{Subset 3:} $\{S\in\Cset^{(m)}: a \in S, b\notin S \}$. Following a very similar procedure to the 2nd subset, we obtain:
\begin{equation*}
    \begin{aligned}
        &\sum_{S\in \Cset^{(m)}: a\notin S, b\in S}\kl(\Pr^a(Y_S^1|S\in\hat\Cset^{(m,1)})\rVert \Pr^b(Y_S^1|S\in\hat\Cset^{(m,1)}))\\
        &=\sum_{S'\in \Cset^{(m-1)}: a, b\notin S'} \bigg[\bigg( \sum_{x\in S'} \frac{w^a(x)}{w^a(S'+b)} \log\frac{w^a(S'+a)}{w^a(S'+b)}\bigg) + \frac{v}{w^a(S'+b)}\log \frac{v}{v+\delta}\cdot\frac{w^a(S'+a)}{w^a(S'+b)} \bigg] \,.
    \end{aligned}
\end{equation*}

Focusing on the terms from the 2nd and 3rd subset and grouping them in an intelligent way gives us
\begin{equation*}
    \begin{aligned}
        &\sum_{S\in \Cset^{(m)}: a\notin S, b\in S}\kl(\Pr^a(Y_S^1|S\in\hat\Cset^{(m,1)})\rVert \Pr^b(Y_S^1|S\in\hat\Cset^{(m,1)}))\\ &+ \sum_{S\in \Cset^{(m)}: a\in S, b\notin S}\kl(\Pr^a(Y_S^1|S\in\hat\Cset^{(m,1)})\rVert \Pr^b(Y_S^1|S\in\hat\Cset^{(m,1)}))\\
        &=\sum_{S'\in \Cset^{(m-1)}: a, b\notin S} \bigg[\bigg( \sum_{x\in S'} \frac{w^a(x)}{w^a(S'+a)} \log\frac{w^a(S'+b)}{w^a(S'+a)} \bigg)+ \frac{v+\delta}{w^a(S'+a)}\log \frac{v+\delta}{v}\cdot\frac{w^a(S'+b)}{w^a(S'+a)} \bigg]\\
        &+\sum_{S'\in \Cset^{(m-1)}: a, b\notin S} \bigg[\bigg( \sum_{x\in S'} \frac{w^a(x)}{w^a(S'+b)} \log\frac{w^a(S'+a)}{w^a(S'+b)}\bigg) + \frac{v}{w^a(S'+b)}\log \frac{v}{v+\delta}\cdot\frac{w^a(S'+a)}{w^a(S'+b)} \bigg] \\
        &=\sum_{S'\in \Cset^{(m-1)}: a, b\notin S} \bigg[\bigg(\frac{v+\delta}{w^a(S'+a)} - \frac{v}{w^a(S'+b)}\bigg)\log \frac{v+\delta}{v} \bigg] \\
        &+\sum_{S'\in \Cset^{(m-1)}: a, b\notin S} \log\frac{w^a(S'+a)}{w^a(S'+b)}\cdot \bigg[\sum_{x\in S'} \bigg(\frac{w^a(x)}{w^a(S'+b)} - \frac{w^a(x)}{w^a(S'+a)}\bigg) +  \frac{v}{w^a(S'+b)} - \frac{v+\delta}{w^a(S'+a)}  \bigg] \\
        &=\sum_{S'\in \Cset^{(m-1)}: a, b\notin S} \bigg[\bigg(\frac{v+\delta}{w^a(S'+a)} - \frac{v}{w^a(S'+b)}\bigg)\log \frac{v+\delta}{v} \bigg]\,.
    \end{aligned}
\end{equation*}
The last equality comes from recognizing that $\sum_{x\in S'} \frac{w^a(x)}{w^a(S'+b)} +\frac{v}{w^a(S'+b)} = \frac{w^a(S'+b)}{w^a(S'+b)} = 1$ and similarly $\sum_{x\in S'} \frac{w^a(x)}{w^a(S'+a)} +\frac{v+\delta}{w^a(S'+a)} = \frac{w^a(S'+a)}{w^a(S'+a)} = 1$ so they cancel out and the second term becomes $0$. We now have the following much more compact identity.
$$ \kl(\Pr^a(\{Y_S^r\})\rVert\Pr^b(\{Y_S^r\})) $$
$$=pR\cdot \log\frac{v+\delta}{v} \cdot \bigg(\sum_{S\in \Cset^{(m)}: a, b \in S} \frac{\delta}{w^a(S)} + \sum_{S'\in \Cset^{(m-1)}: a, b\notin S} \bigg[\frac{v+\delta}{w^a(S'+a)} - \frac{v}{w^a(S'+b)}\bigg] \bigg) \,.$$
To make a connection between this and $\Delta_K^{(m)}$, recognize that under $\calM^a$, $\Delta_K^{(m)} = \tau_a^{(m)} - \tau_b^{(m)}$. We therefore have
\begin{equation*}
\begin{aligned}
\Delta_K^{(m)} &=\tau_a^{(m)} - \tau_b^{(m)}\\
&= \frac{1}{{n-1\choose m-1}} \cdot \bigg( \sum_{S\in \Cset^{(m)}: a\in S} \rho^a(a|S) - \sum_{S\in\Cset^{(m)}:b\in S}\rho^a(b|S)  \bigg)\\
&= \frac{1}{{n-1\choose m-1}} \cdot \bigg( \sum_{S\in \Cset^{(m)}: a\in S} \frac{v+\delta}{w^a(S)}- \sum_{S\in\Cset^{(m)}:b\in S}\frac{v}{w^a(S)}  \bigg)\\
&= \frac{1}{{n-1\choose m-1}} \cdot \bigg( \sum_{S\in \Cset^{(m)}: a, b\in S} \frac{v}{w^a(S)} + \sum_{S\in\Cset^{(m)}:a \in S, b\notin S}\frac{v+\delta}{w^a(S)} - \sum_{S\in\Cset^{(m)}:a \notin S, b\in S}\frac{v}{w^a(S)}  \bigg)\\
&= \frac{1}{{n-1\choose m-1}} \cdot \bigg(\sum_{S\in \Cset^{(m)}: a, b \in S} \frac{\delta}{w^a(S)} + \sum_{S'\in \Cset^{(m-1)}: a, b\notin S'} \bigg[\frac{v+\delta}{w^a(S'+a)} - \frac{v}{w^a(S'+b)} \bigg]\bigg)\,.
\end{aligned}
\end{equation*}
We thus have
\begin{equation}\label{eqn:kl-compact}
    \kl(\Pr^a(\{Y_S^r\})\rVert\Pr^b(\{Y_S^r\})) = pR\log\frac{v+\delta}{v} {n-1\choose m-1} \Delta_K^{(m)} \,.
\end{equation}

We will now bound $\log \frac{v+\delta}{v}$ in terms of $\frac{\Delta^{(m)}_K}{\tau^{(m)}_{K}}$.
\begin{equation*}
\begin{aligned}
\frac{\Delta^{(m)}_K}{\tau_{K}^{(m)}} &= \frac{\sum_{S\in\Cset^{(m)}:a,b\in S} \frac{w^a(a) - w^a(b)}{w^a(S) + 2v+\delta} + \sum_{S\in\Cset^{(m-1)}:a,b\notin S} \frac{w^a(a)}{w^a(S)+v+\delta}-\frac{w^a(b)}{w^a(S)+v}}{\sum_{S\in\Cset^{(m)}:a,b\in S} \frac{w^a(a)}{w^a(S) + 2v+\delta} + \sum_{S\in\Cset^{(m-1)}:a,b\notin S} \frac{w^a(a)}{w^a(S)+v+\delta} }\\
&= \frac{\sum_{S\in\Cset^{(m)}:a,b\in S} \frac{\delta}{w^a(S) + 2v+\delta} + \sum_{S\in\Cset^{(m-1)}:a,b\notin S} \frac{\delta(w^a(S))}{(w^a(S)+v+\delta)(w^a(S)+v)} }{\sum_{S\in\Cset^{(m)}:a,b\in S} \frac{v +\delta}{w^a(S) + 2v+\delta} + \sum_{S\in\Cset^{(m-1)}:a,b\notin S} \frac{v+\delta}{w^a(S)+v+\delta} }\\
&= \frac{\sum_{S\in\Cset^{(m)}:a,b\in S} \frac{1}{w^a(S) + 2v+\delta} + \sum_{S\in\Cset^{(m-1)}:a,b\notin S} \frac{(w^a(S))}{(w^a(S)+v+\delta)(w^a(S)+v)} }{\sum_{S\in\Cset^{(m)}:a,b\in S} \frac{1}{w^a(S) + 2v+\delta} + \sum_{S\in\Cset^{(m-1)}:a,b\notin S} \frac{1}{w^a(S)+v+\delta} } \cdot \frac{\delta}{v+\delta} \\
&\geq \frac{\sum_{S\in\Cset^{(m-1)}:a,b\notin S} \frac{1}{w^a(S)+v+\delta}\cdot \frac{w^a(S)}{w^a(S)+v} }{\sum_{S\in\Cset^{(m-1)}:a,b\notin S} \frac{1}{w^a(S)+v+\delta} } \cdot \frac{\delta}{v+\delta}\\
&\geq \frac{\sum_{S\in\Cset^{(m-1)}:a,b\notin S} \frac{1}{w^a(S)+v+\delta}\cdot \frac{(m-1)v}{(m-1)v + v } }{\sum_{S\in\Cset^{(m-1)}:a,b\notin S} \frac{1}{w^a(S)+v+\delta} } \cdot \frac{\delta}{v+\delta}\\
&= \frac{(m-1)v}{mv }  \cdot \frac{\delta}{v+\delta}\\
&= \frac{m-1}{m} \frac{\delta}{v+\delta} \geq \frac{1}{2}\frac{\delta}{v+\delta}\,.
\end{aligned}
\end{equation*}
We can now bound $\log \frac{v+\delta}{v}$ as
$$ \log \frac{v+\delta}{v} \leq \frac{\delta}{v} \leq \frac{2\delta}{v+\delta} \leq \frac{4\Delta^{(m)}_K}{\tau^{(m)}_{K}} \,.$$
Combining with equation (\ref{eqn:kl-compact}) completes the proof.
\end{proof}

\subsubsection{Approximate top-$K$ recovery}

To obtain a lower bound on the sample complexity for approximate top-$K$ recovery, we require a more nuanced construction than that in the proof of Theorem \ref{thm:lower-bound-all}. At a high level, we need to construct a multiset of $K$-size subsets $\{\Ss^1,\ldots, \Ss^L\}$. Each $K$-size subset $\Ss^l$ corresponds to an MNL model whose top $K$ items are exactly $\Ss^l$. We need to carefully design this multiset such that the pairwise edit distance between any two subsets is larger than $2h$. This automatically ensures that any top-$K$ estimate is close in edit distance to at most one of the $L$ subsets. At the same time, we also want the distance between the distributions induced by any two models to be small in KL divergence sense. This would then allow us to invoke Fano's lemma to lower bound the probability that any statistical estimator outputs a top-$K$ estimate set with small edit distance error. Lastly, we also want $L$ to be sufficiently large in order to obtain a good lower bound using Fano's lemma.

We first present a reworded version of Lemma 9 of \cite{shah2017simple}. The original lemma, in turn, is based on a result due to \cite{levenshtein1971upper} for fixed weighted binary codes.
\begin{lemma}\label{lem:packing-result-shah}\cite[lemma 9]{shah2017simple} Consider the regime where $h <\frac{2}{3}\min\{K, \sqrt n, n-K \}$. For sufficiently large $n$, there exists a multiset $\{s^1,\ldots, s^L\}$ with cardinality $L \geq \exp{\frac{9}{20}h\log n}$ and $s^l \subseteq [\frac{n}{2}, n]$ for $l \in [L]$ such that
$$ |s^l| = \frac{3h}{2}\,\, \forall l\in [L] \quad\text{and}\quad D_{01}(s^j, s^l) = 2h+1 \,\, \text{for all $j\neq l \in [L]$} \,.$$
\end{lemma}

By the above lemma, there exists a multiset of cardinality $\exp{\frac{9}{20}h\log n}$, consisting of sets of size exactly $\frac{3h}{2}$, with elements from $[\frac{n}{2}, n]$. For $s^l$, let us construct a MNL model as follows: let $u = \{1,\ldots, K-\frac{3h}{2}\}$ and $\Ss^l_K = u \cup s^l$ be the set of the top $K$ items for model $l\in [L]$. Note that this construction is valid since $h < \frac{2K}{3}$. Following closely the description of the special MNL model used in subsection (\ref{subsub:proof-exact-top-k}), we assign $v$ as weight for all items in $[n]\backslash \Ss^l_K$, $v+\delta/2$ for all items in $u$ and $v+\delta$ for all items in $s_l$  for some $v > \delta > 0$. In other words, $v^l$ is the set of the top $\frac{3h}{2}$ items for the $l$-th model while $u$ is the set composing of the $\frac{3h}{2}+1$-th, $\ldots, K$-th best item for model $l$.   

We emphasize that by design, $D_{01}(\Ss^l_K, \Ss^j_K) > 2h$ for any $l\neq j\in[L]$. This also means that for any estimate $\hat\Ss_K$, there exists at most one index $a \in [L]$ such that $D_{01}(\hat\Ss_K, \Ss_K^a) \leq h$. In other words, outputting an estimate with low edit distance error is equivalent to exactly identifying the set of the top $K$ items. All that remains is to prove that it is hard to distinguish between any two models as their distributions over observed choice data have small KL divergence. Such a result is summarized in the lemma below:

\begin{lemma}\label{lem:kl-approximate-recover} Consider the construction described above, for any two models indexed by $a,b \in [L]$ with $a\neq b$. Let $\Pr^a$ and $\Pr^b$ be the distributions parametrized by $\Ss^a_K$ and $\Ss_K^b$ as described above, respectively. Under the sampling model described in Section \ref{subsect:sampling-model}, we have
$$ \kl(\Pr^a(\{Y\}) \rVert \Pr^b(\{Y\})) \leq 12hpR{n-1\choose m-1}\frac{{\Delta_{K,h}^{(m)}}^2}{\tau^{(m)}_{K-h}} $$
\end{lemma}

Before proving Lemma \ref{lem:kl-approximate-recover}, let us state and prove the sample complexity lower bound for approximate top-$K$ ranking, the proof of which directly makes use of the lemma.

\begin{theorem}\label{thm:lower-bound-approximate-recovery}
Consider the sampling model described in Section \ref{subsect:sampling-model}.
There exists a choice model $\calM^*$ within the class of Multinomial Logit Models (MNLs) such that for $n \geq 20$, if $pR{n\choose m} \leq \frac{1}{96}\frac{n\log n}{m{\Delta_{K,h}^{(m)}}} \cdot \frac{\tau^{(m)}_{K-h}}{\Delta^{(m)}_{K,h}}$ then any statistical estimator outputs an estimate $\hat\Ss_K$ and
$$ D_{01}(\hat\Ss_K, \Ss^*_K) > h $$
with probability at least $\frac{1}{5}$.
\end{theorem}

\begin{proof} Consider the regime when $p,R$ are small enough such that
$$ pR{n-1\choose m-1} \leq \frac{\log n}{96 {\Delta_{K,h}^{(m)}}^2  }\cdot \tau^{(m)}_{K-h}= \frac{\log n}{96 {\Delta_{K,h}^{(m)}}^2  }\cdot (\tau^{(m)}_{K-h}+\Delta_{K,h}^{(m)}) $$
and the multiset construction as described above. Invoking Fano's inequality, we have the probability that any statistical estimator failing to output an estimate $\hat\Ss_K$ such that $D_{01}(\hat\Ss_K, \Ss^*_K) \leq h$, is lower bounded as
$$ 1-\frac{ \frac{1}{8}h\log n +\log 2 }{\frac{9}{20}h\log n} \geq \frac{1}{5} \quad\text{for } n \geq 20 \,.$$
This finishes the proof.
\end{proof}

We will now proceed to prove Lemma \ref{lem:kl-approximate-recover}.
\begin{proof}
Following the same notation and argument as in the proof of Lemma \ref{lem:kl-div}, we have
$$ \kl(\Pr^a(\{Y\})\rVert \Pr^b(\{Y\})) = pR\sum_{S\in \Cset^{(m)}} \kl(\Pr^a(y_S^1|S\in \hat\Cset^{(m,1)} ) \rVert \Pr^b(y_S^1|S\in \hat\Cset^{(m,1)}))\,. $$
Similarly to the proof of Lemma \ref{lem:kl-div}, let us define some shorthand notations: let $w^a(S)$ denote the sum of the weights of the items in menu $S$ and $w^a(i)$ the weight of item $i$ under the special MNL model whose top $K$ items is $\Ss^a_K$ (similarly defined for $b\neq a$). Expanding on the sum of the KL divergence terms and rearranging the summation order gives us

\begin{equation*}
\begin{aligned}
&\frac{1}{pR}\kl(\Pr^a(\{Y\}) \rVert \Pr^b(\{Y\})) \\
&= \sum_{i\in \Ss_K^{a}\backslash \Ss_K^b} \sum_{S\in C_i^{(m)}} \frac{w^a(i)}{w^a(S)} \log \frac{w^a(i)\cdot w^b(S)}{w^b(i)\cdot w^a(S) } + \sum_{i\in \Ss_K^b \backslash \Ss_K^a }\sum_{S\in C_i^{(m)}} \frac{w^a(i)}{w^a(S)} \log\frac{w^a(i)\cdot w^b(S)}{w^b(i) \cdot w^a(S)} \\
&+ \sum_{i\in \Ss^a_K \cap \Ss_K^b } \sum_{S\in C_i^{(m)}} \frac{w^a(i)}{w^a(S)} \log\frac{w^a(i)w^b(S)}{w^b(i)w^a(S)} +   \sum_{i\in [n] \backslash (\Ss^a_K \cup \Ss_K^b) } \sum_{S\in C_i^{(m)}} \frac{w^a(i)}{w^a(S)} \log\frac{w^a(i)w^b(S)}{w^b(i)w^a(S)}\\
&= \sum_{i\in \Ss_K^{a}\backslash \Ss_K^b} \sum_{S\in C_i^{(m)}} \frac{v+\delta}{w^a(S)} \log \frac{(v+\delta)\cdot w^b(S)}{v\cdot w^a(S) } + \sum_{i\in \Ss_K^b \backslash \Ss_K^a }\sum_{S\in C_i^{(m)}} \frac{v}{w^a(S)} \log\frac{v\cdot w^b(S)}{(v+\delta) \cdot w^a(S)} \\
&+ \sum_{i\in \Ss^a_K \cap \Ss_K^b } \sum_{S\in C_i^{(m)}} \frac{w^a(i)}{w^a(S)} \log\frac{w^b(S)}{w^a(S)} +   \sum_{i\in [n] \backslash (\Ss^a_K \cup \Ss_K^b) } \sum_{S\in C_i^{(m)}} \frac{w^a(i)}{w^a(S)} \log\frac{w^b(S)}{w^a(S)}\\
&= \sum_{i\in \Ss_K^{a}\backslash \Ss_K^b} \sum_{S\in C_i^{(m)}} \frac{v+\delta}{w^a(S)} \log \frac{v+\delta}{v} + \sum_{i\in \Ss_K^b \backslash \Ss_K^a }\sum_{S\in C_i^{(m)}} \frac{v}{w^a(S)} \log\frac{v}{v+\delta} \\
&+\sum_{i\in \Ss_K^{a}\backslash \Ss_K^b} \sum_{S\in C_i^{(m)}} \frac{v+\delta}{w^a(S)} \log \frac{w^b(S)}{w^a(S) } + \sum_{i\in \Ss_K^b \backslash \Ss_K^a }\sum_{S\in C_i^{(m)}} \frac{v}{w^a(S)} \log\frac{w^b(S)}{w^a(S)}\quad\quad(*) \\
&+ \sum_{i\in \Ss^a_K \cap \Ss_K^b } \sum_{S\in C_i^{(m)}} \frac{w^a(i)}{w^a(S)} \log\frac{w^b(S)}{w^a(S)} +   \sum_{i\in [n] \backslash (\Ss^a_K \cup \Ss_K^b) } \sum_{S\in C_i^{(m)}} \frac{w^a(i)}{w^a(S)} \log\frac{w^b(S)}{w^a(S)} \,.
\end{aligned}
\end{equation*}
In the second equality, we have made use of the fact that for any item $i$ in $\Ss_K^a\cup \Ss_K^b$ and $\Ss_K^a \cap \Ss_K^b$, $w^a(i) = w^b(i)$.
Focusing on the first two summations,
\begin{equation*}
\begin{aligned}
&\sum_{i\in \Ss_K^{a}\backslash \Ss_K^b} \sum_{S\in C_i^{(m)}} \frac{v+\delta}{w^a(S)} \log \frac{v+\delta}{v} + \sum_{i\in \Ss_K^b \backslash \Ss_K^a }\sum_{S\in C_i^{(m)}} \frac{v}{w^a(S)} \log\frac{v}{v+\delta} \\
&= \bigg(\sum_{i\in \Ss_K^{a}\backslash \Ss_K^b} \sum_{S\in C_i^{(m)}} \frac{v+\delta}{w^a(S)} - \sum_{i\in \Ss_K^b \backslash \Ss_K^a }\sum_{S\in C_i^{(m)}} \frac{v}{w^a(S)} \bigg) \log \frac{v+\delta}{v} \\
&= (2h+1){n-1\choose m-1} \Delta_{K,h}^{(m)} \log \frac{\delta+v}{v} < 3h{n-1\choose m-1}\Delta_{K,h}^{(m)} \log\frac{\delta+v}{v}\,.
\end{aligned}
\end{equation*}
The last equality comes from recognizing that $|\Ss_K^a \backslash \Ss_K^b| = |\Ss_K^b\backslash \Ss_K^a| = D_{01}(\Ss_K^a, \Ss_K^b) = 2h+1$. We now simplify the remaining four summation terms of $(*)$.
\begin{equation*}
\begin{aligned}
&\sum_{i\in \Ss_K^{a}\backslash \Ss_K^b} \sum_{S\in C_i^{(m)}} \frac{v+\delta}{w^a(S)} \log \frac{w^b(S)}{w^a(S) } + \sum_{i\in \Ss_K^b \backslash \Ss_K^a }\sum_{S\in C_i^{(m)}} \frac{v}{w^a(S)} \log\frac{w^b(S)}{w^a(S)} \\
&+ \sum_{i\in \Ss^a_K \cap \Ss_K^b } \sum_{S\in C_i^{(m)}} \frac{w^a(i)}{w^a(S)} \log\frac{w^b(S)}{w^a(S)} +   \sum_{i\in [n] \backslash (\Ss^a_K \cup \Ss_K^b) } \sum_{S\in C_i^{(m)}} \frac{w^a(i)}{w^a(S)} \log\frac{w^b(S)}{w^a(S)}\\
&= \sum_{i\in [n]} \sum_{S\in \Cset_i^{(m)}} \frac{w^a(i)}{w^a(S)}\log \frac{w^b(S)}{w^a(S)}\\
&= \sum_{S\in \Cset^{(m)}}\sum_{i\in S}\frac{w^a(i)}{w^a(S)}\log \frac{w^b(S)}{w^a(S)}\\
&= \sum_{S\in\Cset^{(m)}} \log \frac{w^b(S)}{w^a(S)}\\
&= \log \frac{\prod_{S\in \Cset^{(m)}} w^b(S)  }{ \prod_{S\in\Cset^{(m)}} w^a(S)  }\\
&= \log 1 = 0\,.
\end{aligned}
\end{equation*}
The last equality comes from recognizing that the two models $a$ and $b$ only differ by the the identities of the items while the weights are the same. Therefore, the two products in the log term are identical. We thus obtain
$$\kl(\Pr^a(\{Y\}) \rVert \Pr^b(\{Y\})) < pR3h{n-1\choose m-1} \Delta_{K,h}^{(m)} \log\frac{\delta+v}{v}\,. \quad(**)$$
Following the same argument as in the proof of Lemma \ref{lem:kl-div}, we can bound
$$\log \frac{\delta+v}{v} \leq 4\frac{\Delta_{K,h}^{(m)}}{\tau_{K-h}^{(m)}} \,.$$
Substituting this into $(**)$ completes the proof.
\end{proof}

In short, the (matching) lower bound on the sample complexity for approximate top-$K$ ranking is
$$ \Omega\bigg(\frac{n\log n}{m{\Delta_{K,h}^{(m)}}} \cdot \big( 1+ \frac{\tau^{(m)}_{K+h+1}}{\Delta^{(m)}_{K,h}}) \bigg) \,.$$

%% file: proofs/proofs_menu_size.tex
\subsection{Effect of the menu size on the sample complexity}

\begin{reptheorem}{thm:mnl-variational-lower-bound} Consider an MNL model with $n\geq 2$ items and fix a $K$, we have
    \begin{equation*}
     \frac{1}{m\Delta_K^{(m)}} = \theta\bigg( \frac{1}{e^{U_K}-e^{U_{K+1}}}\cdot \big(1 + \frac{1}{m-1}\big)^2  \bigg)\,,
    \end{equation*}
    \begin{equation*}
    \frac{\tau^{(m)}_{K+1}}{\Delta_K^{(m)}} = \theta\bigg( \frac{e^{U_{K+1}}}{e^{U_K}-e^{U_{K+1}}}\cdot \big( 1+ \frac{1}{m-1}\big) \bigg) \,.
    \end{equation*}
\end{reptheorem}

\begin{proof}

\textbf{Part 1}. We will prove that there exists a positive valued function $l(m)$ such that $m\Delta_K^{(m)} < l(m)$ and that $l(m)$ monotonically increases with $m$, and at diminishing rate. Once we have obtained $l(m)$, we can set $f_1(m) = \frac{1}{l(m)}$. In fact, we will prove a slightly more general result concerning any pair of items $i, j$ where $w_i := e^{U_i} > e^{U_j} =: w_j$. Let $\Delta^{(m)}_{ij} = \tau_i^{(m)}- \tau_j^{(m)}$. Additionally, let $w_n := e^{U_n} = e^{U_{\min}}$ and $w_1 := e^{U_1} = e^{U_{\max}}$ be two model-dependent constants. We have
\begin{equation*}
\begin{aligned}
m\Delta^{(m)}_{ij} &= \frac{m}{{n-1\choose m-1}} \cdot \bigg(\sum_{S\in \Cset^{(m)}_i \cap \Cset^{(m)}_j} \frac{w_i}{w_i + w_j + \sum_{k\in S} w_k} -\frac{w_j}{w_i + w_j + \sum_{k\in S}w_k} \\
&+ \sum_{S' \in \Cset^{(m-1)}: i, j\notin S} \frac{w_i}{w_i + \sum_{k\in S} w_k } - \frac{w_j}{w_j + \sum_{k\in S} w_k}   \bigg)\\
&= \frac{m(w_i -w_j)}{{n-1\choose m-1}} \bigg( \sum_{S\in \Cset^{(m)}_i \cap \Cset^{(m)}_j} \frac{1}{w_i + w_j + \sum_{k\in S} w_k} + \sum_{S' \in \Cset^{(m-1)}: i, j\notin S} \frac{1}{w_j + \sum_{k\in S} w_k}\cdot  \frac{\sum_{k\in S} w_k}{w_i + \sum_{k\in S} w_k}  \bigg)\\
&\leq \frac{m(w_i -w_j)}{{n-1\choose m-1}} \bigg( {n-2\choose m-2} \cdot \frac{1}{w_i + w_j + (m-2)w_n} + {n-2\choose m-1} \frac{1}{w_j + (m-1)w_n} \cdot  \frac{(m-1)w_1}{w_i + (m-1)w_1}  \bigg)\\
&=  \frac{m(w_i -w_j)}{n-1} \bigg( (m-1) \cdot \frac{1}{w_i + w_j + (m-2)w_n} + (n-m)\frac{1}{w_j + (m-1)w_n} \cdot  \frac{(m-1)w_1}{w_i + (m-1)w_1}  \bigg)\\
&= \frac{(w_i -w_j)}{n-1} \bigg(n\frac{m}{[w_j + (m-1)w_n]}\frac{(m-1)w_1}{[w_i + (m-1)w_1]} \\
&+ m(m-1)\cdot \bigg[ \frac{1}{w_i + w_j + (m-2)w_n} - \frac{mw_1}{[w_i + (m-1)w_1][w_j + (m-1)w_n]} \bigg]  \bigg)\\
&= \frac{(w_i -w_j)}{n-1} \bigg(n\frac{m}{[mw_n + w_j -w_n]}\frac{(m-1)w_1}{[w_i + (m-1)w_1]} \\
&+ m(m-1)\cdot \frac{(w_1-w_i)(w_n - w_j)}{[(m-2)w_n+w_i+w_j][(m-1)w_n + w_j][(m-1)w_1+w_i]}  \bigg)\\
&= \frac{(w_i -w_j)}{n-1} \bigg(n\frac{m}{[mw_n + w_j -w_n]}\frac{(m-1)w_1}{[w_i + (m-1)w_1]} \\
&+ \underbrace{\frac{(w_1-w_i)(w_n - w_j)}{(m-2)w_n+w_i+w_j}\cdot \frac{m}{mw_n + w_j -w_n} \cdot\frac{(m-1)}{(m-1)w_1+w_i]}}_{\leq 0}  \bigg)\\
&\leq (w_i -w_j)\cdot \frac{n}{n-1} \cdot \frac{m}{[mw_n + w_j -w_n]}\frac{(m-1)w_1}{[w_i + (m-1)w_1]}\\
&=(w_i -w_j)\cdot \frac{n}{n-1} \cdot (1 - \frac{w_j/w_n-1}{m-1 + w_j/w_n}) \cdot (1-\frac{w_i/w_1}{m-1+w_i/w_1}) \,.
\end{aligned} 
\end{equation*}

It is easy to see that all of the terms containing $m$ are positive and \emph{increase} with $m$. Additionally, it is easy to show that all of these terms increase with \emph{decreasing} rate with respect to $m$.
Though complicated, the above lower bound can somewhat be simplified using big-$O$ notations.
$$ m\Delta^{(m)}_{ij} = \Omega\bigg( (w_i-w_j) \cdot (1-\frac{1}{m})^2 \bigg)\,. $$
By modifying the above argument, we can also show a variational lower bound on $m\Delta^{(m)}_{ij}$.

\begin{equation*}
\allowdisplaybreaks
\begin{aligned}
m\Delta^{(m)}_{ij} &= \frac{m}{{n-1\choose m-1}} \cdot \bigg(\sum_{S\in \Cset^{(m)}_i \cap \Cset^{(m)}_j} \frac{w_i}{w_i + w_j + \sum_{k\in S} w_k} -\frac{w_j}{w_i + w_j + \sum_{k\in S}w_k} \\
&+ \sum_{S' \in \Cset^{(m-1)}: i, j\notin S} \frac{w_i}{w_i + \sum_{k\in S} w_k } - \frac{w_j}{w_j + \sum_{k\in S} w_k}   \bigg)\\
&= \frac{m}{{n-1\choose m-1}} \cdot \bigg(\sum_{S\in \Cset^{(m)}_i \cap \Cset^{(m)}_j} \frac{w_i - w_j}{w_i + w_j + \sum_{k\in S} w_k} + \sum_{S' \in \Cset^{(m-1)}: i, j\notin S} \frac{w_i}{w_i + \sum_{k\in S} w_k } - \frac{w_j}{w_j + \sum_{k\in S} w_k}   \bigg)\\
&= \frac{m}{{n-1\choose m-1}} \cdot \bigg(\sum_{S\in \Cset^{(m)}_i \cap \Cset^{(m)}_j} \frac{w_i - w_j}{w_i + w_j + \sum_{k\in S} w_k} + \sum_{S' \in \Cset^{(m-1)}: i, j\notin S} \frac{(w_i-w_j)(\sum_{k\in S} w_k)}{(w_i + \sum_{k\in S} w_k)(w_j + \sum_{k\in S} w_k)}   \bigg)\\
&= \frac{m(w_i -w_j)}{{n-1\choose m-1}} \bigg( \sum_{S\in \Cset^{(m)}_i \cap \Cset^{(m)}_j} \frac{1}{w_i + w_j + \sum_{k\in S} w_k} + \sum_{S' \in \Cset^{(m-1)}: i, j\notin S} \frac{1}{w_i + \sum_{k\in S} w_k}\cdot  \frac{\sum_{k\in S} w_k}{w_j + \sum_{k\in S} w_k}  \bigg)\\
&[\text{Using the fact that } \frac{\sum_{k\in S} w_k }{w_j + \sum_{k\in S} w_k} \geq \frac{\sum_{k\in S} w_n }{w_j + \sum_{k\in S} w_n} \,\forall S: j \in S] \\
&\geq \frac{m(w_i -w_j)}{{n-1\choose m-1}} \bigg( \sum_{S\in \Cset^{(m)}_i \cap \Cset^{(m)}_j} \frac{1}{w_i + w_j + \sum_{k\in S} w_k} + \sum_{S' \in \Cset^{(m-1)}: i, j\notin S} \frac{1}{w_i + \sum_{k\in S} w_k}\cdot  \frac{(m-1) w_n}{w_j + (m-1) w_n}  \bigg)\\
&\geq \frac{m(w_i -w_j)}{{n-1\choose m-1}} \bigg( {n-2\choose m-2} \frac{1}{w_i + w_j + (m-2) \bar w} + {n-2\choose m-1} \frac{1}{w_i + (m-1)\bar w}\cdot  \frac{(m-1) w_n}{w_j + (m-1) w_n}  \bigg)\,.
\end{aligned}
\end{equation*}

where $\bar w = \frac{1}{n-2}\sum_{k\neq i, j} w_k$. The last inequality comes from applying Jensen's inequality. More precisely, one could treat the sum $\sum_{k\in S} w_k$ as a random variable and the summation as the unnormalized `expectation' over the uniform distribution of the menus. As $\frac{1}{x}$ is a convex function in $x$ for positive $x$, Jensen's inequality applies. Furthermore, without loss of generality, one can assume that $\bar w = 1$ (via scaling of the weights). Consequently, $w_n \leq 1$. Continuing with the expansion gives us

\begin{equation*}
\allowdisplaybreaks
\begin{aligned}
m\Delta^{(m)}_{ij} &\geq \frac{m(w_i -w_j)}{{n-1\choose m-1}} \bigg( {n-2\choose m-2} \frac{1}{w_i + w_j + (m-2) \bar w} + {n-2\choose m-1} \frac{1}{w_i + (m-1)\bar w}\cdot  \frac{(m-1) w_n}{w_j + (m-1) w_n}  \bigg)\\
&=  \frac{m(w_i -w_j)}{n-1} \bigg(  \frac{m-1}{w_i + w_j + (m-2)} + \frac{n-m}{w_i + (m-1)}\cdot  \frac{(m-1) w_n}{w_j + (m-1) w_n}  \bigg)\\
&= \frac{(w_i -w_j)}{n-1} \bigg(  \frac{m(m-1)}{w_i + w_j + (m-2) } + \frac{m(n-m)}{w_i + (m-1)}\cdot  \frac{(m-1) w_n}{w_j + (m-1) w_n}  \bigg)\\
&\geq \frac{(w_i -w_j)}{n-1} \bigg(  \frac{m(m-1)}{w_i + w_j + (m-2)} + \frac{(m-1)(n-m)}{w_i + (m-1)}\cdot  \frac{(m-1) w_n}{w_j + (m-1) w_n}  \bigg)\\
&\geq \frac{(w_i -w_j)}{n-1} \bigg( n\cdot \frac{m-1}{m-1+w_i} \cdot \frac{(m-1)w_n}{(m-1)w_n+w_j} + \frac{m(m-1)}{w_i + w_j + (m-2)} - \frac{m(m-1)(m-1)w_n}{(w_i + m-1)(w_j + (m-1)w_n)}   \bigg)\\
&\geq \frac{(w_i -w_j)}{n-1} \bigg( n\cdot \frac{m-1}{m-1+w_i} \cdot \frac{(m-1)w_n}{(m-1)w_n+w_j} \\
&+m(m-1)\bigg( \frac{1}{w_i + w_j + (m-2)} - \frac{(m-1)w_n}{(w_i + m-1)(w_j + (m-1)w_n)} \bigg)  \bigg)\\
&[\text{Expanding on the difference term}]\\
&\geq \frac{(w_i -w_j)}{n-1} \bigg( n\cdot \frac{m-1}{m-1+w_i} \cdot \frac{(m-1)w_n}{(m-1)w_n+w_j} \\
&+m(m-1)\bigg( \frac{1}{w_i + w_j + (m-1)} - \frac{(m-1)w_n}{(w_i + m-1)(w_j + (m-1)w_n)} \bigg)  \bigg)\\
&= \frac{(w_i -w_j)}{n-1} \bigg( n\cdot \frac{m-1}{m-1+w_i} \cdot \frac{(m-1)w_n}{(m-1)w_n+w_j} \\
&+m(m-1)\cdot (w_iw_j + w_i(m-1)w_n + mw_j +m(m-1)w_n - w_j - (m-1)w_n - w_i(m-1)w_n - w_j(m-1)w_n - \\
&m(m-1)w_n+1(m-1)w_n))/\big((w_i + w_j + (m-1))(w_i + m-1)(w_j + (m-1)w_n)\big) \bigg)\\
&= \frac{(w_i -w_j)}{n-1} \bigg( n\cdot \frac{m-1}{m-1+w_i} \cdot \frac{(m-1)w_n}{(m-1)w_n+w_j} \\
&+m(m-1)\cdot \frac{mw_j + w_iw_j - w_j - w_j(m-1)w_n}{(w_i + w_j + (m-1))(w_i + m-1)(w_j + (m-1)w_n)}  \bigg)\\
&= \frac{(w_i -w_j)}{n-1} \bigg( n\cdot \frac{m-1}{m-1+w_i} \cdot \frac{(m-1)w_n}{(m-1)w_n+w_j} \\
&+m(m-1)w_j\cdot \frac{m + w_i - 1 - (m-1)w_n}{(w_i + w_j + (m-1))(w_i + m-1)(w_j + (m-1)w_n)}  \bigg)\\
&= \frac{(w_i -w_j)}{n-1} \bigg( n\cdot \frac{m-1}{m-1+w_i} \cdot \frac{(m-1)w_n}{(m-1)w_n+w_j} + \frac{m(m-1)w_j((m-1)(1-w_n) + w_i)}{(w_i + w_j + (m-1))(w_i + m-1)(w_j + (m-1)w_n)}  \bigg)\\
&\geq  \frac{(w_i -w_j)}{n-1} \bigg( n\cdot \frac{m-1}{m-1+w_i} \cdot \frac{(m-1)w_n}{(m-1)w_n+w_j} + \frac{m(m-1)w_j((m-1)(1-w_n) + w_i)}{(w_i + w_j + m)(w_i + m-1)(w_j + (m-1)w_n)}  \bigg)\\
&= \frac{(w_i -w_j)}{n-1} \bigg( n\cdot \frac{m-1}{m-1+w_i} \cdot \frac{(m-1)w_n}{(m-1)w_n+w_j} + \underbrace{w_j\cdot \frac{m}{w_i+w_j+m}\cdot \frac{m-1}{w_j + (m-1)w_n}\cdot \frac{w_i + (m-1)(1-w_n)}{w_i+(m-1)}}_{\geq 0} \bigg)\\
&\geq \frac{(w_i -w_j)}{n-1} \bigg( n\cdot (1-\frac{w_i}{m-1+w_i}) \cdot (1-\frac{w_j}{(m-1)w_n+w_j}) \bigg) \\
&=(w_i -w_j)\cdot \frac{n}{n-1} \cdot (1-\frac{w_i}{m-1+w_i}) \cdot (1-\frac{w_j/w_n}{m-1+w_j/w_n})  \,.
\end{aligned}
\end{equation*}

We thus have $m\Delta_{ij} = O((w_i - w_j) \cdot 1-\frac{1}{m})^2)$. Combining with the upper bound shown earlier, we get
$$m\Delta_{ij} = \theta\bigg((w_i - w_j) \cdot (1-\frac{1}{m})^2\bigg) \,.$$

\textbf{Part 2.} We will prove that there exists a positive valued function $l(m)$ such that $\frac{\Delta^{(m)}_{ij}}{\tau^{(m)}_{j}} \geq l(m)$ and that $l(m_2) > l(m_1)$ for $m_2 > m_1 \geq 2$. Consider for any two items $i, j$ such that $w_i > w_j$. By definition
\begin{equation*}
\begin{aligned}
\frac{\Delta^{(m)}_{ij}}{\tau_{j}^{(m)}} &= \frac{\sum_{S\in \Cset_i^{(m)} \cap \Cset_j^{(m)}} \frac{w_i - w_j}{\sum_{k\in S} w_k } +  \sum_{S\in \Cset^{(m-1)}: i,j \notin S} \frac{w_i}{w_i + \sum_{k\in S} w_k } -  \frac{w_j}{w_j + \sum_{k\in S} w_k }  }{ \sum_{S\in \Cset_i^{(m)} \cap \Cset_j^{(m)}} \frac{w_j}{\sum_{k\in S} w_k } +  \sum_{S\in \Cset^{(m-1)}: i,j \notin S} \frac{w_j}{w_j + \sum_{k\in S} w_k } }\\
&= \frac{\sum_{S\in \Cset_i^{(m)} \cap \Cset_j^{(m)}} \frac{w_i - w_j}{\sum_{k\in S} w_k } +  \sum_{S\in \Cset^{(m-1)}: i,j \notin S} \frac{(w_i-w_j)(\sum_{k\in S}w_k )}{(w_i + \sum_{k\in S} w_k)(w_j + \sum_{k\in S} w_k) }  }{ \sum_{S\in \Cset_i^{(m)} \cap \Cset_j^{(m)}} \frac{w_j}{\sum_{k\in S} w_k } +  \sum_{S\in \Cset^{(m-1)}: i,j \notin S} \frac{w_j}{w_j + \sum_{k\in S} w_k } }\\
&= \frac{\sum_{S\in \Cset_i^{(m)} \cap \Cset_j^{(m)}} \frac{1}{\sum_{k\in S} w_k } +  \sum_{S\in \Cset^{(m-1)}: i,j \notin S} \frac{\sum_{k\in S}w_k }{(w_i + \sum_{k\in S} w_k)(w_j + \sum_{k\in S} w_k) }  }{ \sum_{S\in \Cset_i^{(m)} \cap \Cset_j^{(m)}} \frac{1}{\sum_{k\in S} w_k } +  \sum_{S\in \Cset^{(m-1)}: i,j \notin S} \frac{1}{w_j + \sum_{k\in S} w_k } } \cdot \frac{w_i - w_j}{w_j}\\
&= (1- \frac{\sum_{S\in \Cset^{(m-1)}: i,j \notin S} \frac{w_j }{(w_i + \sum_{k\in S} w_k)(w_j + \sum_{k\in S} w_k) }}{ \sum_{S\in \Cset_i^{(m)} \cap \Cset_j^{(m)}} \frac{1}{\sum_{k\in S} w_k } +  \sum_{S\in \Cset^{(m-1)}: i,j \notin S} \frac{1}{w_j + \sum_{k\in S} w_k }}) \cdot \frac{w_i - w_j}{w_j}\,.
\end{aligned}
\end{equation*}

The goal is to upper (and lower) bound the quantity
$$ \frac{\sum_{S\in \Cset^{(m-1)}: i,j \notin S} \frac{w_j }{(w_i + \sum_{k\in S} w_k)(w_j + \sum_{k\in S} w_k) }}{ \sum_{S\in \Cset_i^{(m)} \cap \Cset_j^{(m)}} \frac{1}{\sum_{k\in S} w_k } +  \sum_{S\in \Cset^{(m-1)}: i,j \notin S} \frac{1}{w_j + \sum_{k\in S} w_k }} \,.$$
For a lower bound, we can show that
\begin{equation*}
\begin{aligned}
&\frac{\sum_{S\in \Cset^{(m-1)}: i,j \notin S} \frac{w_j }{(w_i + \sum_{k\in S} w_k)(w_j + \sum_{k\in S} w_k) }}{ \sum_{S\in \Cset_i^{(m)} \cap \Cset_j^{(m)}} \frac{1}{\sum_{k\in S} w_k } +  \sum_{S\in \Cset^{(m-1)}: i,j \notin S} \frac{1}{w_j + \sum_{k\in S} w_k }}\\
&\geq \frac{{n-2\choose m-1} \frac{1}{(m-1)w_1+w_i} \frac{w_j}{w_j + (m-1)w_1} }{{n-2\choose m-2}\cdot \frac{1}{(m-2)w_n + w_i + w_j}  + {n-2\choose m-1}\cdot \frac{1}{(m-1)w_n + w_j}   }\\
&= \frac{ (n-m)\cdot \frac{1}{(m-1)w_1+w_i} \frac{w_j}{w_j + (m-1)w_1} }{(m-1)\cdot \frac{1}{(m-2)w_n + w_i + w_j}  + (n-m)\cdot \frac{1}{(m-1)w_n + w_j}   }\\
&[\text{Noting that } (m-2)w_n + w_i + w_j = (m-1)w_n + w_j + (w_i - w_n) \geq (m-1)w_n + w_j ]\\
&\geq \frac{ (n-m)\cdot \frac{1}{(m-1)w_1+w_i} \frac{w_j}{w_j + (m-1)w_1} }{(m-1)\cdot \frac{1}{(m-1)w_n + w_j}  + (n-m)\cdot \frac{1}{(m-1)w_n + w_j}   }\\
&= \frac{n-m}{n-1}\cdot \frac{1}{(m-1)w_1+w_i} \cdot \frac{(m-1)w_n + w_j}{(m-1)w_1+w_j}\\
&\geq \frac{n-m}{n-1}\cdot \frac{1/w_1}{m-1+w_i/w_1} \cdot \frac{w_n}{w_1}\,.
\end{aligned}
\end{equation*}
For an upper bound, we can show that
\begin{equation*}
\begin{aligned}
&\frac{\sum_{S\in \Cset^{(m-1)}: i,j \notin S} \frac{w_j }{(w_i + \sum_{k\in S} w_k)(w_j + \sum_{k\in S} w_k) }}{ \sum_{S\in \Cset_i^{(m)} \cap \Cset_j^{(m)}} \frac{1}{\sum_{k\in S} w_k } +  \sum_{S\in \Cset^{(m-1)}: i,j \notin S} \frac{1}{w_j + \sum_{k\in S} w_k }}\\
&\leq \frac{{n-2\choose m-1} \frac{1}{(m-1)w_n+w_i} \frac{w_j}{w_j + (m-1)w_n} }{{n-2\choose m-2}\cdot \frac{1}{(m-2)w_1 + w_i + w_j}  + {n-2\choose m-1}\cdot \frac{1}{(m-1)w_1 + w_j}   }\\
&= \frac{ (n-m)\cdot \frac{1}{(m-1)w_n+w_i} \frac{w_j}{w_j + (m-1)w_n} }{(m-1)\cdot \frac{1}{(m-2)w_1 + w_i + w_j}  + (n-m)\cdot \frac{1}{(m-1)w_1 + w_j}   }\\
&[\text{Noting that } (m-2)w_1 + w_i + w_j = (m-1)w_1 + w_j + (w_i - w_1) \leq (m-1)w_1 + w_j ]\\
&\leq \frac{ (n-m)\cdot \frac{1}{(m-1)w_n+w_i} \frac{w_j}{w_j + (m-1)w_n} }{(m-1)\cdot \frac{1}{(m-1)w_1 + w_j}  + (n-m)\cdot \frac{1}{(m-1)w_1 + w_j}   }\\
&= \frac{n-m}{n-1}\cdot \frac{1/w_n}{m-1+w_i/w_n} \cdot \frac{w_1}{w_n}\,.
\end{aligned}
\end{equation*}
In both cases, it is straight forward to check that both of the upper and lower bounds decrease with increasing $m$ and at a decreasing rate. Furthermore, they have the same asympotic dependency on $m$. Putting both bounds together, we have
$$ \frac{\Delta^{(m)}_{ij}}{\tau^{(m)}_{j}} = \theta\bigg( (1-\frac{n-m}{n-1}\cdot \frac{1}{m}) \cdot \frac{w_i - w_j}{w_j} \bigg) = \theta\bigg( \frac{n}{n-1}\cdot (1-\frac{1}{m}) \cdot \frac{w_i - w_j}{w_j} \bigg) \,. $$
Simplifying the above expression by removing the dependency on $n$ gives
$$\frac{\Delta^{(m)}_{ij}}{\tau^{(m)}_{j}} = \theta\bigg( (1-\frac{1}{m}) \cdot \frac{w_i - w_j}{w_j} \bigg)\,. $$
This completes the proof.
\end{proof}

%% file: proofs/proofs_connection.tex
\subsection{Connection among MNL-MLE, Generalized Borda Count and Spectral Ranking}

In this section, we will prove the theorems about the connections among the three algorithms. Here, we provide more general results pertaining to ranking output (as opposed to just top $K$ estimate).

\begin{reptheorem}{thm:borda-mle-connection} Consider the sampling model described in Section \ref{subsect:sampling-model}, for any $p > 0$, in the limit as $R\rightarrow \infty$, MNL-MLE and choice-based Borda count will produce the same top $K$ estimate. Moreover, this holds even if the data does not come from the MNL model or any IID-RUM.
\end{reptheorem}

\begin{proof} 
\textbf{Notation:} As a shorthand notation, we will use $p(i|S)$ denote the probability of item $i$ being chosen from menu $S$. Note that this probability does not necessarily follow any parametric RUM. Let $\text{ord}$ denote the ordering function whose input is a set of real numbers and whose output is the full ordering of the indices of those numbers in increasing order.
Under our sampling model, in the limit as $R\rightarrow \infty$, all the possible menus in $\Cset^{(m)}$ are observed given that $p > 0$. Furthermore, given infinite data, the observed probability becomes exact. Consider the log likelihood function
$$ \Loss(\mb U) =\sum_{i=1}^n \bigg(\sum_{S\in \Cset_i^{(m)}} p(i|S) \cdot \log \frac{e^{U_i}}{\sum_{j\in S} e^{U_j}}\bigg) \,.$$

The derivative of the log likelihood with respect to individual partworth parameter is given as
\begin{equation*}
\begin{aligned}
\nabla_{U_i}\Loss(\mb U) &=  \sum_{S\in \Cset_i^{(m)}} \bigg( -\sum_{j\in S, j\neq i} p(j|S) \frac{e^{U_i}}{\sum_{k\in S} e^{U_k}} -  p(i|S) \frac{e^{U_i}}{\sum_{k\in S} e^{U_k}} + p(i|S) \bigg) \\
&= \sum_{S\in \Cset_i^{(m)}} \bigg( p(i|S) - \frac{e^{U_i}}{\sum_{k\in S} e^{U_k} }   \bigg)\,.
\end{aligned}
\end{equation*}
As the log likelihood function is concave in $U$, the MLE estimate $\hat {\mb U}$ must satisfy:
$$ \sum_{S\in \Cset_i^{(m)}} \frac{e^{\hat U_i}}{\sum_{j\in S} e^{\hat U_j}} = \sum_{S\in \Cset_i^{(m)}}  p(i|S) \quad \forall i \,.$$
On the other hand it can easily be shown that for any two items $i\neq j$,
$$ \hat U_i > \hat U_j \Rightarrow \sum_{S\in \Cset_i^{(m)}} \frac{e^{\hat U_i}}{\sum_{k\in S}e^{\hat U_k}} > \sum_{S\in \Cset_j^{(m)}} \frac{e^{\hat U_j}}{\sum_{k\in S}e^{\hat U_k}} \,.$$
Hence, 
$$ \text{ord}(\{\hat U_i\}_{i=1}^n) = \text{ord}(\{\sum_{S\in \Cset_i^{(m)}}\frac{e^{\hat U_i}}{\sum_{k\in S} e^{\hat U_k} } \}_{i=1}^n ) = \text{ord}(\{\sum_{S\in \Cset_i^{(m)}} p(i|S)\}_{i=1}^n ) \,. $$
Observing that the ordering given by the Borda Count algorithm is consistent with the ordering induced by $\{\sum_{S\in \Cset_i^{(m)}} p(i|S)\}_{i=1}^n$ completes the proof.
\end{proof} 

\begin{reptheorem}{thm:borda-mle-asr-connection} Consider the sampling model described in Subsection \ref{subsect:sampling-model}. Assume that the underlying choice model generating the data is in the class of IID-RUMs whose noise distribution has absolutely continuous density function with support on the real line. For any $p > 0$, in the limit as $R\rightarrow \infty$, then Spectral Ranking, MNL-MLE and choice-based Borda count produce the same top $K$ estimate.

On the other hand, there exists a choice model where in the limit as $R\rightarrow \infty$, the spectral ranking algorithm produces a different top $K$ estimate from MNL-MLE/Borda count.
\end{reptheorem}

\begin{proof} Following the argument as in the proof of Theorem \ref{thm:borda-mle-connection}, given $p > 0$ as $R\rightarrow \infty$, all possible menus of size $m$ are observed and all choice probabilities are exact. For the first part of the theorem it suffices to prove that under IID-RUMs, the Spectral Ranking algorithm is consistent in recovering the true ordering among the items, as this will be the same ordering as given by Generalized Borda Count/MLE.

The Spectral Ranking algorithm ranks the items by the stationary distribution of a Markov Chain constructed using choice data. For analysis purpose, we follow the construction due to \cite{negahban2017rank,maystre2015fast}. Fix a menu size $m \geq 2$ and consider the following Markov Chain where for any two items $i, j$,
$$ \M_{ij} =\begin{cases}\frac{1}{{n-1\choose m-1}} \sum_{S\in\Cset^{(m)}: i, j\in S} \rho(j|S)&\text{if } j\neq i \\
1 - \sum_{k\neq i} \M_{ik} &\text{if } j = i
\end{cases} \,.$$
Consider any two items $i, j$ such that $U_i > U_j$. Under IID-RUMs assumptions in the theorem statement, we have
$$ \rho(i|S) > \rho(j|S) \quad\forall S \in\Cset^{(m)}: i, j \in S $$
and
$$ \rho(i|S' \cup \{i\}) > \rho(j | S' \cup \{j\}) \quad \forall S' \in \Cset^{(m-1)}: i, j\notin S \,.$$
It is easy to show that
$$ \M_{ki} > \M_{kj} \quad\forall k\neq i, j $$
and
$$ \M_{ii} > \M_{jj}\,. $$
Let $\pi$ denote the stationary distribution of the Markov Chain constructed above.
By definition, the stationary distribution of the Markov Chain satisfies
\begin{equation*}
\begin{aligned}
\pi_i &= \sum_{k\in [n]} \pi_k\cdot \M_{ki} > \sum_{k\in [n]} \pi_k \cdot \M_{kj} = \pi_j \,.
\end{aligned}
\end{equation*}

For the second half of the theorem, let us consider the pairwise comparison setting ($m=2$). We will construct a pairwise choice model such that MNL-MLE/Generalized Borda Count give a different ordering among the items from Spectral Ranking algorithm in the limit as $R\rightarrow \infty$. Consider an universe of 4 items with the following pairwise choice probability. Note that $P_{ij} = P(j|\{i, j\})$. Define
$$P = \begin{bmatrix}
0.5 &0.6 &0.55 &0.55\\
0.2 &0.5 &0.85 &0.60\\
0.45 &0.40 &0.5 &0.95\\
0.45 &0.45 &0.15 &0.5\\
\end{bmatrix} \,.$$
It is easy to check that $\tau_4 > \tau_3 > \tau_2 > \tau_1$. In the limit of infinite data, MLE/Borda Count will output the ordering $4,3,2,1$ (best item first). However, the Spectral Ranking algorithm will give the ordering $4,2,3,1$. This completes the proof.\end{proof}

%% file: experiments/experiments_long.tex
\section{Additional experiments}\label{sect:experiment}

\subsection{Dataset descriptions}

Table (\ref{tab:dataset}) shows the characteristics of the datasets used in our experiments: number of items, number of rankings available and whether the data contains partial rankings.
\begin{table}[H]
\centering
\begin{tabular}{|l|l|l|l|}
\hline
\multicolumn{1}{|c|}{Dataset} & \multicolumn{1}{c|}{$n$} & \multicolumn{1}{c|}{Num rankings} & \multicolumn{1}{c|}{Partial ranking?} \\ \hline
APA  &5      &$ 6000$  &No                                       \\ \hline
Sushi  &11      &$ 5000$   &No                                       \\ \hline
Irish-North   &12   &$ 44k$   &Yes   \\ \hline
Irish-West &9 &$ 30k$ & Yes \\ \hline
Irish-Meath &14 &$ 64k$  & Yes  \\ \hline
F1 &22 &18 &Yes                                       \\ \hline
\end{tabular}
\caption{\label{tab:dataset}Characteristics of the datasets used in our experiments.}
\end{table}

\textbf{Data preparation:} Given menu size $m$, to estimate choice probability from rankings and partial rankings, we first enumerate all the possible menus of size $m$. For each menu $S$, and for each $i\in S$, we count the number of rankings where $i$ is ranked ahead of all the other items in $S$, a 'win' for $i$. For a partial ranking where there are $l$ items ranked as `equal', we count $\frac{1}{l}$-`win' for each item. Choice probability is obtained by normalizing the number of wins by the number of rankings.

\textbf{Experiments: } For each independent trial, we first generate all the data as specified by the maximum sample size. We then feed increasing portion of this data to the algorithm and check if the algorithm correctly identifies \emph{all} of the top $K$ items. For all experiments, we keep $R=100$ and adjust $p$ to obtain the appropriate expected sample size.

\subsection{Additional experimental results}

\begin{figure}[H]
\centering
    \includegraphics[scale=0.5]{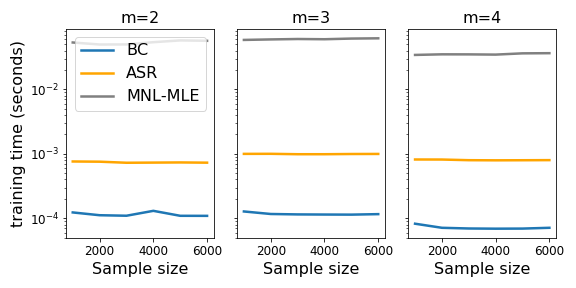}
\caption{
    \textbf{APA election dataset:} Average training time (seconds) against sample size for $m=2,3,4$.}
\end{figure}

\begin{figure}[H]
\centering
\includegraphics[scale=0.5]{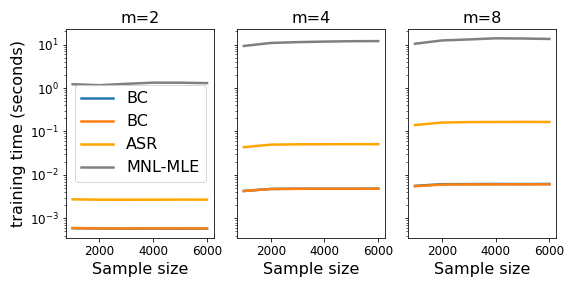}
\caption{\textbf{Irish North dataset:} Average training time (seconds) against sample size for $m=2,4,8$.}
\end{figure}

\begin{figure}[H]
\centering
\includegraphics[scale=0.5]{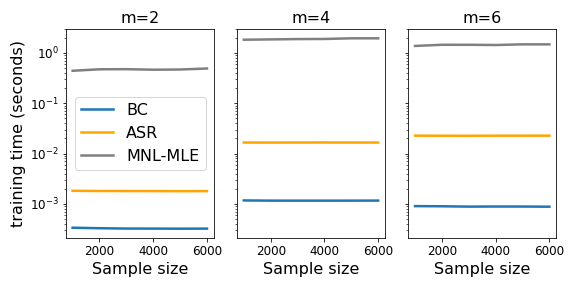}
\caption{\textbf{Irish West dataset:} Average training time (seconds) against sample size for $m=2,4,6$.}
\end{figure}

\begin{figure}[H]
\centering
    \includegraphics[scale=0.5]{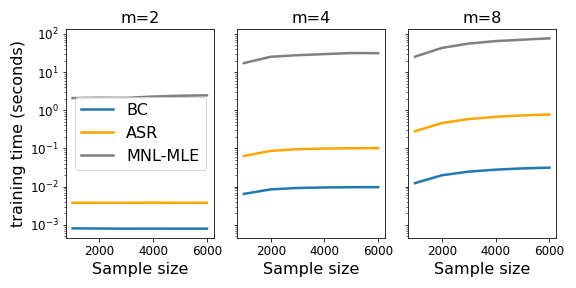}
\caption{\textbf{Irish Meath dataset:} Average training time (seconds) against sample size for $m=2,4,8$.}
\end{figure}
            
\begin{figure}[H]
\centering
    \includegraphics[scale=0.5]{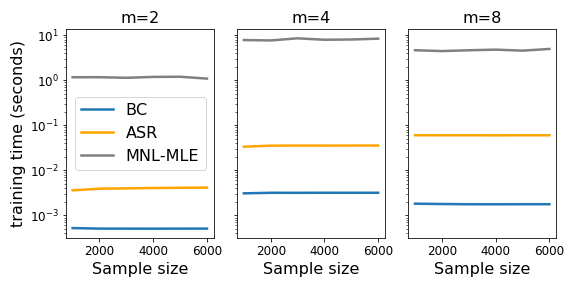}
\caption{\textbf{SUSHI dataset:} Average training time (seconds) against sample size for $m=2,4,8$.}
\end{figure}

\begin{figure}[H]
\centering
    \includegraphics[scale=0.5]{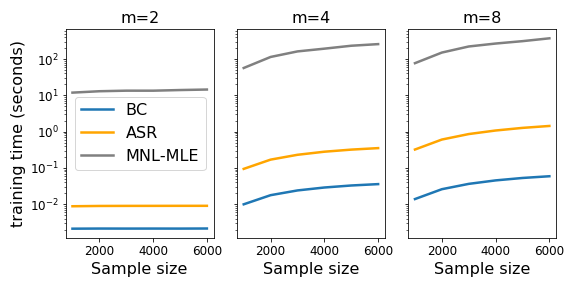}
\caption{\textbf{F1 dataset:} Average training time (seconds) against sample size for $m=2,4,8$. }
\end{figure}

\newpage
\begin{figure}[H]
\centering
    \includegraphics[scale=0.5]{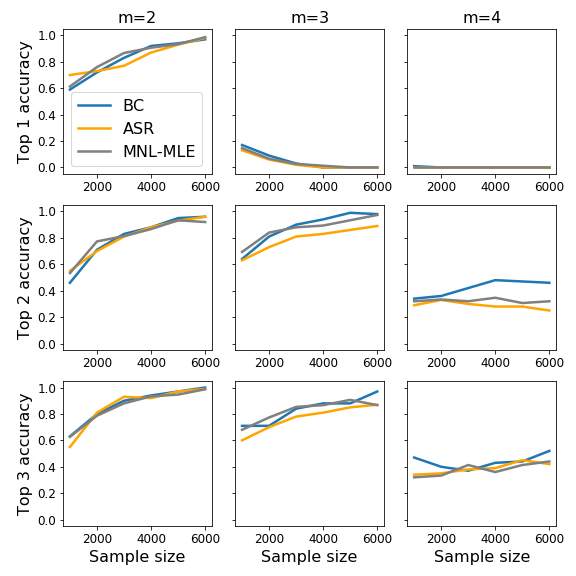}
\caption{
    \textbf{APA election dataset:} Exact top $K$ accuracy against sample size}

\label{fig:apa-experiment}
\end{figure}

\newpage
\begin{figure}[H]
\centering
    \includegraphics[scale=0.5]{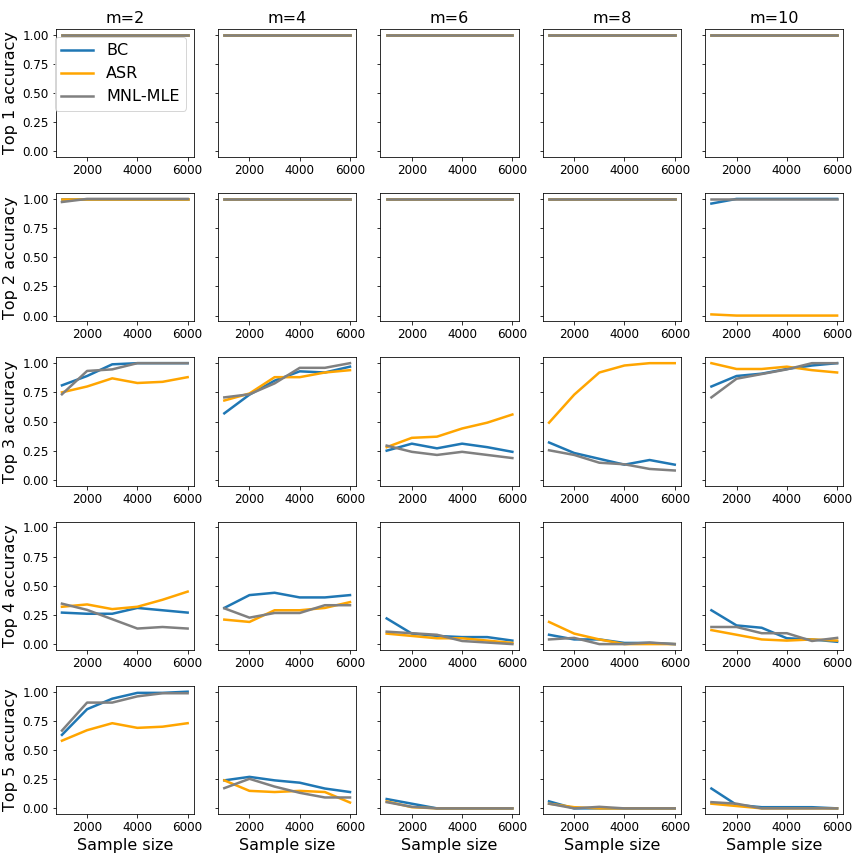}
\caption{\textbf{SUSHI dataset:} Exact top $K$ accuracy against sample size}
\label{fig:sushi-experiment}
\end{figure}

\newpage
\begin{figure}[H]
\centering
    \includegraphics[scale=0.5]{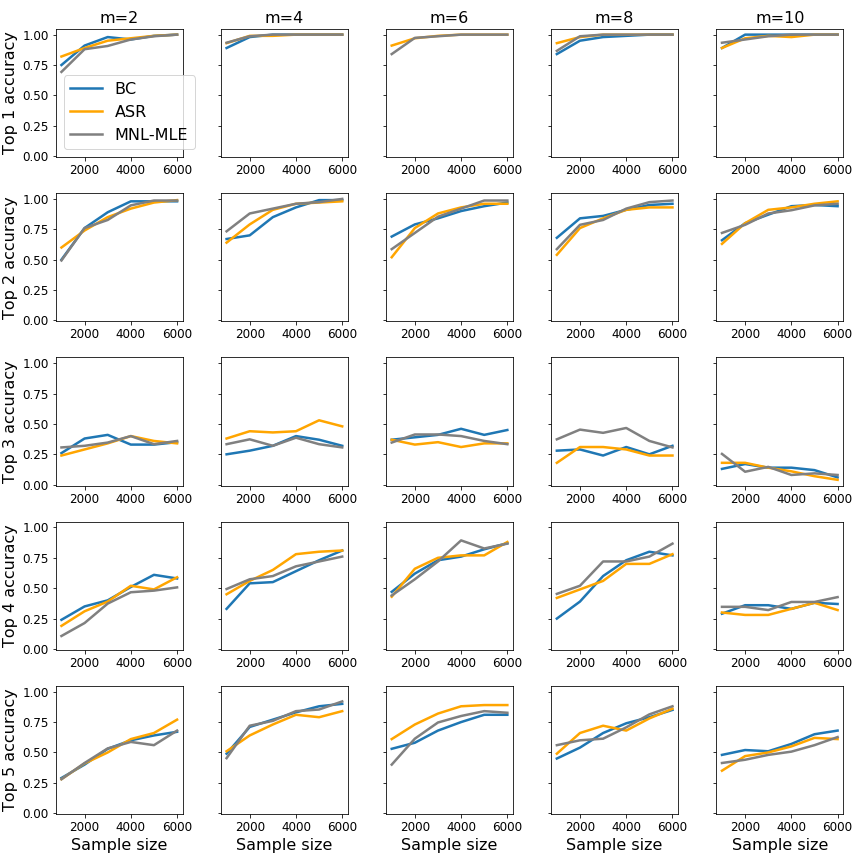}
\caption{\textbf{Irish North dataset:} Exact top $K$ accuracy against sample size}
\label{fig:irish-north-experiment}
\end{figure}

\newpage
\begin{figure}[H]
\centering
    \includegraphics[scale=0.5]{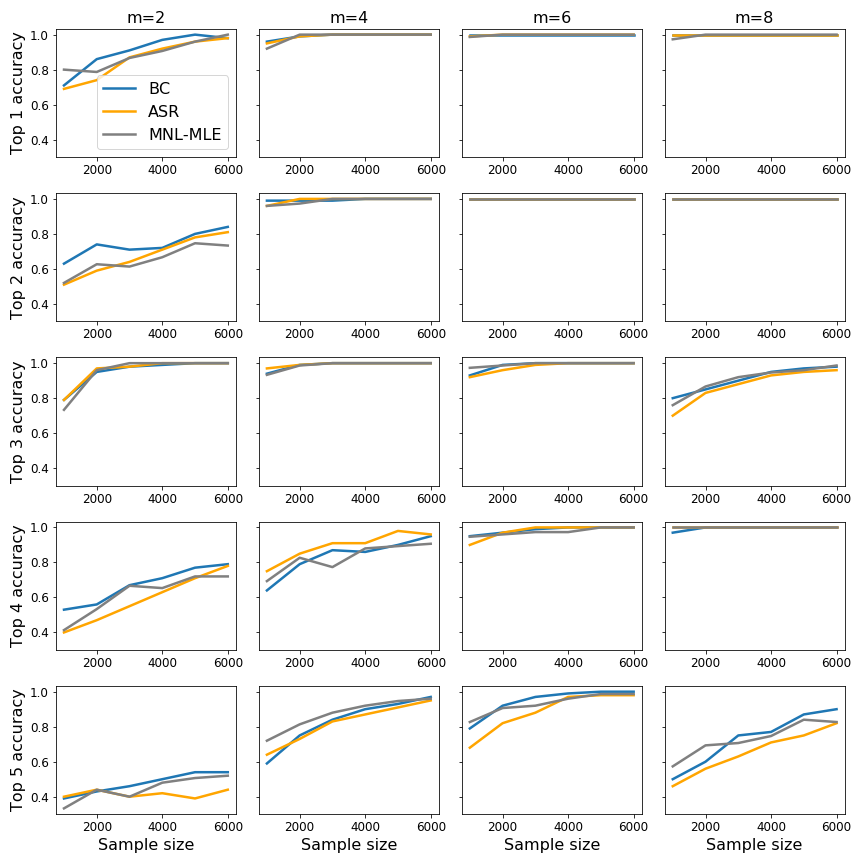}
\caption{\textbf{Irish West dataset:} Exact top $K$ accuracy against sample sizes}
\label{fig:irish-west-experiment}
\end{figure}

\newpage
\begin{figure}[H]
\centering
    \includegraphics[scale=0.5]{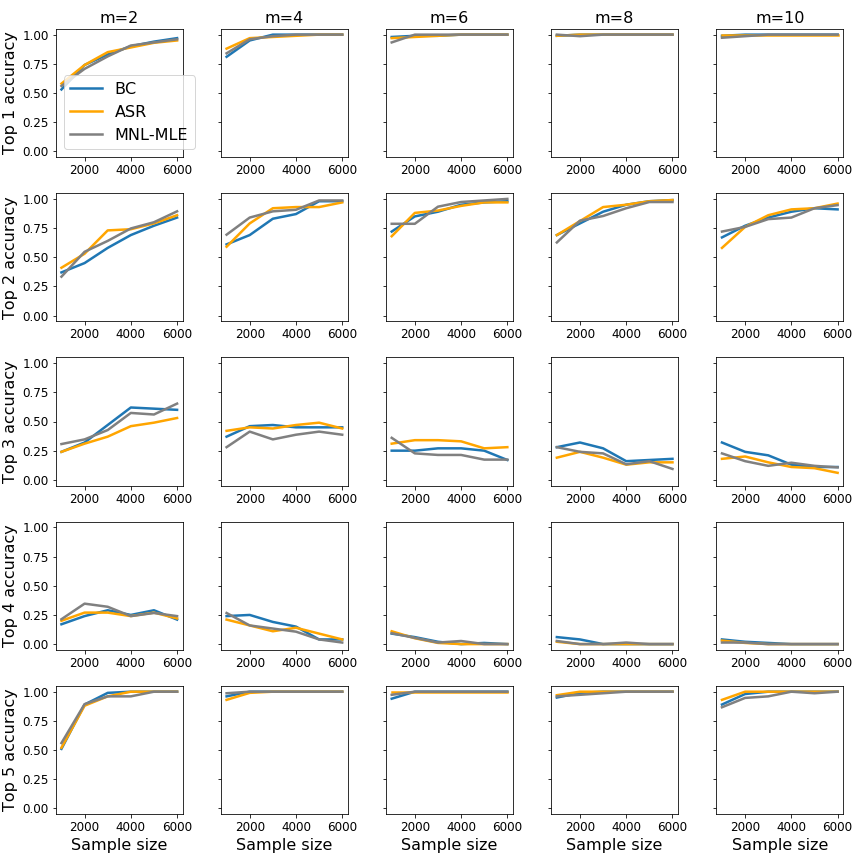}
\caption{\textbf{Irish Meath dataset:} Exact top $K$ accuracy against sample size}
\label{fig:irish-meath-experiment}
\end{figure}

\newpage
\begin{figure}[H]
\centering
    \includegraphics[scale=0.35]{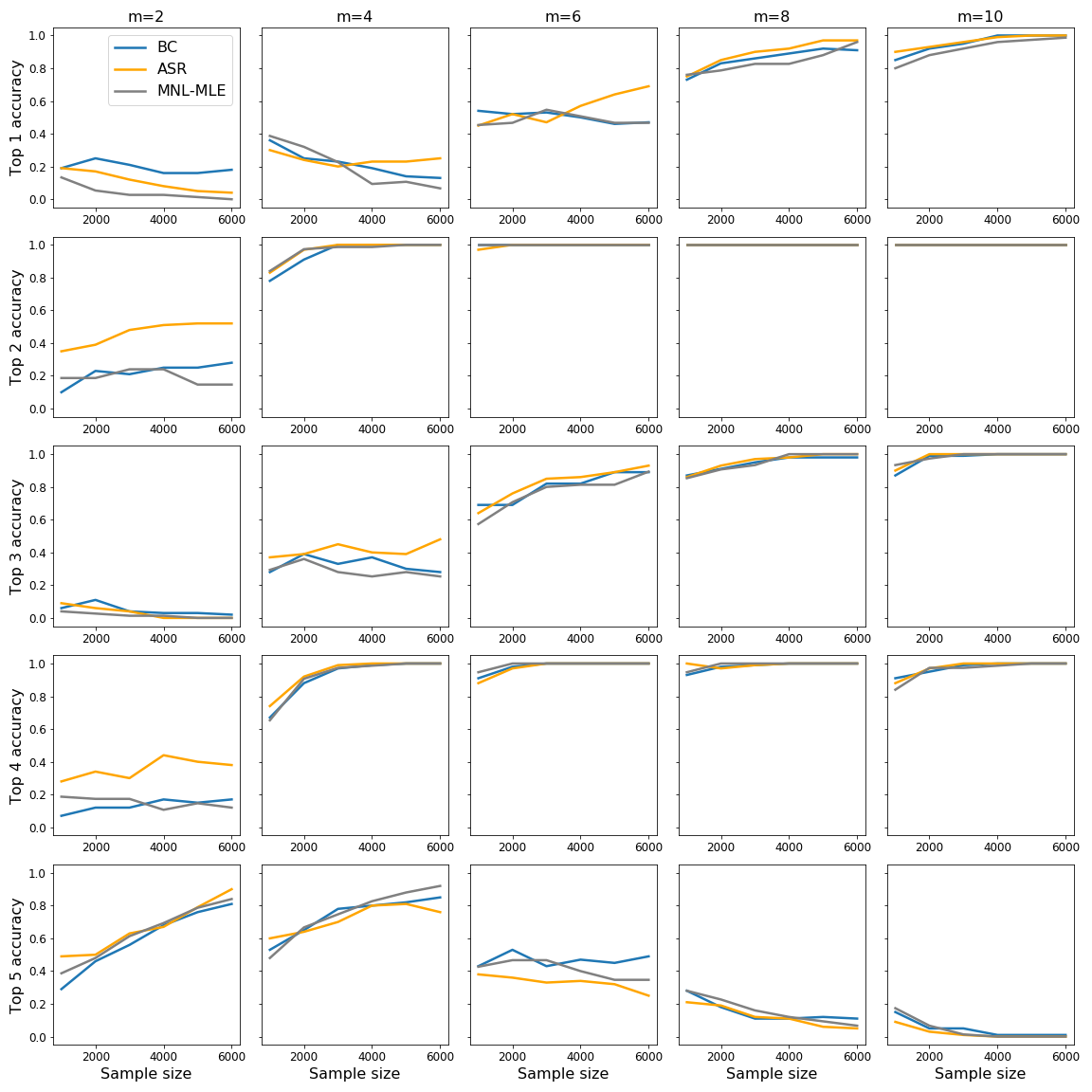}
\caption{\textbf{F1 dataset:} Exact top $K$ accuracy against sample size. }
\label{fig:f1-experiment}
\end{figure}